\documentclass{bmvc2k}

\title{MCDS-VSS: Moving Camera Dynamic Scene Video Semantic Segmentation by Filtering with Self-Supervised Geometry and Motion}

\addauthor{Angel Villar-Corrales}{villar@ais.uni-bonn.de}{1}
\addauthor{Moritz Austermann}{austermann@ais.uni-bonn.de}{1}
\addauthor{Sven Behnke}{behnke@cs.uni-bonn.de}{1}

\addinstitution{
	Autonomous Intelligent Systems – Computer Science Institute VI, \\
	Center for Robotics and \\
	Lamarr Institute for Machine Learning and Artificial Intelligence, \\
	University of Bonn, Germany
}

\runninghead{Villar-Corrales \etal}{MCDS-VSS: Video Semantic Segmentation}

\def\etal{\emph{et al}\bmvaOneDot}

\usepackage{amssymb}
\usepackage{amsmath}
\usepackage{algorithm}% http://ctan.org/pkg/algorithms
\usepackage[noend]{algpseudocode}
\usepackage{nth}
\usepackage[all]{nowidow}
\usepackage{bbm}
\usepackage{booktabs,etoolbox}
\usepackage{multirow}
\usepackage{caption}
\usepackage{graphicx}
\usepackage{subcaption}

\makeatletter
\def\algbackskip{\hskip-\ALG@thistlm}
\makeatother

\usepackage{array}
\usepackage{ragged2e}
\newcolumntype{P}[1]{>{\centering\arraybackslash}p{#1}}
\newcolumntype{R}[1]{>{\RaggedLeft\arraybackslash}p{#1}}

\usepackage[export]{adjustbox}
\usepackage{arydshln}

\usepackage{amsmath}

\newcommand{\StructSeg}{\text{MCDS-VSS} }
\newcommand{\StructSegNosp}{\text{MCDS-VSS}}

\newcommand{\Images}{\mathcal{X}}

\newcommand{\ImageIdx}[1]{\mathbf{x}_{#1}}

\newcommand{\EgoImageIdx}[1]{\mathbf{\hat{x}}^\textrm{ego}_{#1}}

\newcommand{\ImgFeatsIdx}[1]{\mathbf{h}_{#1}}

\newcommand{\FeatsIdx}[1]{\mathbf{s}_{#1}}
\newcommand{\EgoFeatsIdx}[1]{\mathbf{s}^\textrm{ego}_{#1}}
\newcommand{\FlowFeatsIdx}[1]{\mathbf{s}^\textrm{full}_{#1}}

\newcommand{\UpdateGate}{\mathbf{u}}
\newcommand{\Conv}{\text{Conv}}
\newcommand{\bias}{b}

\newcommand{\Segmentations}{\mathcal{Y}}

\newcommand{\SegmentationIdx}[1]{\mathbf{y}_{#1}}

\newcommand{\PredSegmentationIdx}[1]{\mathbf{\hat{y}}_{#1}}

\newcommand{\FilterSegmentationIdx}[1]{\mathbf{\tilde{y}}_{#1}}

\newcommand{\Depth}{\mathbf{d}}
\newcommand{\InvDepth}{\mathbf{d}^{\triangleleft}}
\newcommand{\NormInvDepth}{\widetilde{\mathbf{d}^{\triangleleft}}}
\newcommand{\NormInvDepthIdx}[1]{\widetilde{\mathbf{d}^{\triangleleft}_{#1}}}

\newcommand{\MinDepth}{D_{\mathrm{min}}}
\newcommand{\MaxDepth}{D_{\mathrm{max}}}

\newcommand{\CamTFIdx}[2]{\textrm{C}_{{#1}}^{\,{#2}}}
\newcommand{\CamState}{\mathbf{c}}
\newcommand{\CamStateIdx}[1]{\mathbf{c}_{#1}}

\newcommand{\Intrinsics}{K}

\newcommand{\FeatState}{\mathbf{s}}
\newcommand{\FeatStateIdx}[1]{\mathbf{s}_{#1}}

\newcommand{\FlowIdx}[2]{\textrm{F}_{#1}^{\,#2}}

\newcommand{\PredFlowIdx}[2]{\hat{\textrm{F}}_{{#1}}^{\,{#2}}}

\newcommand{\FlowStateIdx}[1]{\mathbf{s}^{f}_{t}}

\newcommand{\ForwardWarp}{\mathcal{F}_\textrm{fwd}}

\newcommand{\ImageEncoder}{\mathcal{E}_\textrm{x}}
\newcommand{\SegmentationDecoder}{\mathcal{D}_\textrm{y}}
\newcommand{\DepthDecoder}{\mathcal{D}_\textrm{d}}
\newcommand{\MotionEncoder}{\mathcal{E}_\textrm{m}}
\newcommand{\CameraPredictor}{\mathcal{D}_\textrm{c}}

\newcommand{\ResidualFlowEncoder}{\mathcal{R}_\textrm{f}}

\newcommand{\CameraUpdate}{\text{ConvGRU}}

\newcommand{\LossSeg}[2]{\mathcal{L}_\textrm{Segm}({#1},{#2})}
\newcommand{\LossSegEmpty}{\mathcal{L}_\textrm{Segm}}

\newcommand{\LossCamEmpty}{\mathcal{L}_\textrm{Ego}}

\newcommand{\LossDepthEmpty}{\mathcal{L}_\textrm{Depth}}
\newcommand{\LossPhoto}[2]{\mathcal{L}_\textrm{Photo}({#1},{#2})}
\newcommand{\LossPhotoEmpty}{\mathcal{L}_\textrm{Photo}}
\newcommand{\DepthReg}[1]{\mathcal{L}_\textrm{Reg}({#1})}
\newcommand{\DepthRegEmpty}{\mathcal{L}_\textrm{Reg}}
\newcommand{\SSIM}{\text{SSIM}}

\newcommand{\LossFlowEmpty}{\mathcal{L}_\textrm{Flow}}

\newcommand{\LossStructSeg}{\mathcal{L}}
\newcommand{\LossConst}[2]{\mathcal{L}_\textrm{TC}({#1},{#2})}
\newcommand{\LossConstEmpty}{\mathcal{L}_\textrm{TC}}

\usepackage{hyperref}
\usepackage{booktabs}                        % for nicer looking tables
\usepackage{tabularx}
\usepackage{multicol}
\usepackage{tikz}
\def\R{{\mathbb R}}

\newcommand{\Figure}[1]{Figure~\ref{#1}}

\newcommand{\Figures}[2]{Figures~\ref{#1} and \ref{#2}~}
\newcommand{\Figuress}[2]{Figures~\ref{#1}--\ref{#2}}

\newcommand{\Equation}[1]{Equation~\eqref{#1}}

\newcommand{\Table}[1]{Table~\ref{#1}}
\newcommand{\Tables}[2]{Tables~\ref{#1}~and~\ref{#2}}
\newcommand{\Section}[1]{Section~\ref{#1}}

\newcommand{\Appendix}[1]{Appendix~\ref{#1}}

\let\svthefootnote\thefootnote
\newcommand\freefootnote[1]{%
	\let\thefootnote\relax%
	\footnotetext{#1}%
	\let\thefootnote\svthefootnote%
}

\usepackage{hyperref}
\usepackage[dvipsnames]{xcolor}
\usepackage[titletoc]{appendix}

\begin{document}

\maketitle

%%%%%%%%%%%%%%%%%%%%%%%%%%%%%%%%%%%%%%%%%%%%%%%%%%%%%%%%%%%%%%%%%%%%%%%%%%%%%%%%
\begin{abstract}

	Autonomous systems, such as self-driving cars, rely on reliable semantic environment perception for decision making.
	Despite great advances in video semantic segmentation, existing approaches ignore important inductive biases and lack structured and interpretable internal representations.
	In this work, we propose \emph{\StructSegNosp}, a structured filter model that learns in a self-supervised manner to estimate scene geometry and ego-motion of the camera, while also estimating the motion of external objects. Our model leverages these representations to improve the temporal consistency of semantic segmentation without sacrificing segmentation accuracy.
	\StructSeg follows a prediction-fusion approach in which scene geometry and camera motion are first used to compensate for ego-motion, then residual flow is used to compensate motion of dynamic objects, and finally the predicted scene features are fused with the current features to obtain a temporally consistent scene segmentation.
	Our model parses automotive scenes into multiple decoupled interpretable representations such as scene geometry, ego-motion, and object motion.
	Quantitative evaluation shows that \StructSeg achieves superior temporal consistency on video sequences while retaining competitive segmentation performance.
	Code and pretrained models are available in the \href{https://sites.google.com/view/mcds-vss/home}{project website}.
\end{abstract}

%%%%%%%%%%%%%%%%%%%%%%%%%%%%%%%%%%%%%%%%%%%%%%%%%%%%%%%%%%%%%%%%%%%%%%%%%%%%%%%%
\section{Introduction}

Video semantic segmentation (VSS) is the task of assigning a categorical label to each pixel in every frame of a video sequence~\cite{Zhou_SurveyDeepLearningTechniqueVideoSegmentation_2022}.
This task is highly relevant in the field of robotics, where understanding and interpreting scenes from video is crucial for many applications, e.g. autonomous driving~\cite{Siam_ComparativeStudeySegmetationAutonomousDriving_2018} or indoor service tasks~\cite{Seichter_EfficientRGBDSemanticSegmentationForIndoorSceneAnalysis_2021}.
Thanks to the availability of high-quality image datasets, semantic segmentation of automotive scenarios has recently seen tremendous progress~\cite{zhao_PSPNet_2017, Chen_DeepLabV3+_2018, Wang_HRNetVisualRecognition_2020}.
However, obtaining temporally consistent segmentation of video sequences still remains a challenge due to the lack of large-scale annotated video datasets and the lack of suitable inductive biases for video processing.

To address these limitations, existing VSS models enforce temporal continuity by propagating features across multiple frames through the use of unstructured recurrent networks~\cite{Pfeuffer_SemanticSegmentaitonVideoSequencesConvLSTM_2019, Siam_ConvolutionalGatedRecurrentNetworksVideoSegmnetation_2017}, optical flow models~\cite{Ding_EveryFrameCountsLearningVideoSegmentationAndOpticalFlow_2020, Gadde_SemanticVideoCNNsThroughRepresentationWarping_2017}, or transformers~\cite{Wang_TemporalMemoryAttentionVideoSemanticSegmenation_2021, Li_VideoSemanticSegmentationViaSparseTemporalTransformer_2021};
thus exploiting temporal correlations in the video sequences in a data-driven manner.

However, these models ignore specific properties from the target domain, which could potentially be incorporated into the model architecture in order to improve its performance and generalization capabilities.
For instance, in the automotive domain, the observations taken from a moving vehicle can be decomposed into \emph{static background features}, which move only due to the ego-motion of the vehicle, and \emph{dynamic object features} that correspond to moving objects.
Incorporating such motion and geometric inductive biases into the network architecture can lead to models producing a more temporally consistent interpretation of the scene, outperforming models that attempt to learn these properties solely from data.

To test this hypothesis, we propose \emph{\StructSegNosp}, a structured recurrent model that explicitly incorporates geometry and motion inductive biases from the \emph{moving camera dynamic scene} (MCDS) domain in order to improve the temporal consistency of a segmentation network.
\StructSeg follows a prediction-fusion approach in which ego-motion is compensated by projecting scene features into the current time-step using estimated scene geometry and estimated camera motion. Estimated residual flow is then used to compensate for object motion. Finally, the predicted features are fused with the features extracted from the current frame to obtain a temporally consistent semantic segmentation of the scene.

Through self-supervised learning (SSL), \StructSeg learns to estimate scene geometry and ego-motion. It also estimates motion of additional moving objects (e.g. pedestrians or vehicles), and hard-wires our knowledge from the MCDS domain to project the previous scene features into the current time-step using these representations.
The structured design of our filter allows us to factorize the perceived complex changes in the scene into simpler factors of variation; thus easing the modeling of temporal information.

Our experiments show that \StructSeg improves the temporal consistency of a segmentation model without compromising its segmentation performance, outperforming VSS baselines which ignore moving camera dynamic scene inductive biases, and performing comparably to state-of-the-art VSS models.
Furthermore, \StructSeg parses an automotive scene into interpretable internal representations, such as depth, camera motion, and object flow.

\vspace*{1.2mm}
\noindent In summary, our contributions are as follows:
\vspace{-0.2cm}
\begin{itemize}
	\item We propose \StructSegNosp, a structured recurrent filter that improves the temporal consistency of a segmentation model without sacrificing segmentation performance.
	\vspace{-0.16cm}

	\item \StructSeg learns depth and ego-motion in a self-supervised way, and uses these representations together with estimated object motion to propagate scene features.
	\vspace{-0.16cm}

	\item Our model outperforms existing VSS baselines on Cityscapes---achieving superior temporal consistency and parsing the scene into human-interpretable representations.
\end{itemize}

%%%%%%%%%%%%%%%%%%%%%%%%%%%%%%%%%%%%%%%%%%%%%%%%%%%%%%%%%%%%%%%%%%%%%%%%%%%%%%%%
\section{Related Work}

\paragraph{Video Semantic Segmentation:}

VSS methods are often divided into two distinct categories.
The first class aims to reduce the computational cost and improve the efficiency of segmentation models, instead of naively encoding and interpreting every single input frame.
Several methods improve the efficiency by propagating and reusing features extracted from selected key frames~\cite{Jain_AccelEfficientVideoSemanticSegmentation_2019, Zhu_DeepFeatureFlowVideoRecognition_2017};
whereas other approaches achieve efficiency by employing lightweight neural network blocks~\cite{Pfeuffer_SeparableConvolutionalLSTMFasterVideoSegmentation_2019, Shelhamer_ClockworkConvnetsForVideoSemanticSegmentation_2016}
or by distilling the information from large teacher models into smaller models~\cite{Liu_EfficientSemanticVideoSegmentationWithPerFrameInference_2020}.
The second category, to which our proposed method belongs, aims to improve the semantic segmentation performance and temporal consistency by exploiting the temporal continuity of video streams.
Some methods exploit temporal dependencies between video frames and improve the consistency of the predicted segmentation maps by combining image segmentation models with recurrent neural networks (RNNs)~\cite{Siam_ConvolutionalGatedRecurrentNetworksVideoSegmnetation_2017, Pfeuffer_SemanticSegmentaitonVideoSequencesConvLSTM_2019, Shelhamer_ClockworkConvnetsForVideoSemanticSegmentation_2016, Villar_MSPredVideoPrediction_2022} or with attention-based modules~\cite{Wang_TemporalMemoryAttentionVideoSemanticSegmenation_2021, Lao_SimultaneouslyShortAndLongTermTemporalModelingVSS_2023, Sun_LearningLocalAndGlobalTemporalContextsForVideoSemanticSegmentatio_2024, Li_VideoSemanticSegmentationViaSparseTemporalTransformer_2021, Sun_MiningRelationsAcrossCrossFrameAffinitiesVSS_2022}.
Another family of works use an optical flow module to compute the feature correspondence between consecutive frames, and then use this flow for predictive feature learning~\cite{Ding_EveryFrameCountsLearningVideoSegmentationAndOpticalFlow_2020,Zhu_DeepFeatureFlowVideoRecognition_2017, Liu_EfficientSemanticVideoSegmentationWithPerFrameInference_2020, Xu_DynamicVideoSegmentationNetwork_2018, Baghbaderani_TemporallyConsistentVideoSemanticSegmenationBidirectionalOcclusionFeaturePropagation_2024, Varghese_TC_2021, Hur_JointOpticalFlowAndTemporallyConsistentSemSeg_2016}.

Our method belongs to the latter category of VSS models.
However, unlike aforementioned approaches, \StructSeg incorporates assumptions from the domain of moving cameras and dynamic scenes into the model design, and computes interpretable intermediate geometry and motion-aware representations, which lead to accurate and temporally consistent video segmentation results.

\paragraph{Improving Segmentation via Depth~\&~Camera Motion Estimation:}

Self-supervised depth estimation (SSDE) aims to learn the scene geometry from unlabeled monocular videos, without any recorded depth information.
This is often achieved by training a neural network to jointly predict the scene depth and camera ego-motion between two video frames, synthesizing the second frame from the first using differentiable warping, and minimizing a photometric loss function~\cite{Garg_UnuspervisedCNNSingleViewDepthEstimation_2016, Zhou_UnsupervisedLearningDethEgoMotionFromVideo_2017, bian_SCDepth_2021, Gordon_DepthFromVideosInTheWild_2019, Wang_LearningDepthMonoculaVideosDirectMethods_2018}.

The interplay between semantic segmentation and SSDE has been studied for various tasks, including depth estimation~\cite{Zama_GeometryMeetsSemantics_2019}, domain adaptation~\cite{guizilini_geometric_2021,Kuznietsov_TowardsUnsupervisedOnlineDomainAdaptationForSemanticSegmentation_2022,Wang_DomainAdaptiveSemanticSegmentationWithSSDE_2021}, and semi-supervised learning~\cite{hoyer_ThreeWaysToImproveSemanticSegmentation_2021, novosel_BoostingSemanticSegmentationWithMultiTaskSelfSupervisedLearning_2019}.
These models exploit SSDE as an additional source of supervision, helping segmentation models learn high-level semantic features, especially when few labeled samples are available.

Several works have investigated the use of depth and motion for semantic segmentation in videos.
Approaches like~\cite{Shi_SpSequenceNetSegmentation4DPointClouds_2020, Athar_4DFormer_2023, Cao_SLCFNet_2024, Zhu_4DPanopticSegmentationInvariantEquivariantFieldPrediction_2023} segment dynamic scenes by jointly processing video frames with depth information captured by LiDAR scanners;
whereas methods like~\cite{Ma_MultiViewDeepLearningSemSegMappingRGBDCameras_2017, Kundu_VirtualMultiViewFusion3DSemanticSegmentation_2020, Chen_SpatialInformationGuidedConvolutionRGBDSemseg_2021, Rosu_SemisupervisedSegmanticMappingThroughLabelPropagationWithSemanticTexture_2020} use depth and camera pose information
in combination with a semantic segmentation model in order to improve the segmentation performance by enforcing consistency between predictions from multiple viewpoints.
Recently, depth-aware panoptic segmentation models~\cite{Qiao_VIPDeepLab_2021,Petrovai_MonoDVPSdepthAwareVSS_2023, Yuan_Polyphonicformer_2022} aim to jointly solve the tasks of panoptic segmentation and depth estimation by extending a segmentation model with a depth decoder and conditioning its prediction using instance-masks.

The method most similar to ours is Wagner \etal~\cite{Wagner_FunctionallyModularSemanticSegmentation_2018},
which leverages depth and camera motion learned in a supervised manner to improve the performance of a segmentation model on video sequences.
However, this method has several limitations, including not modeling moving objects and requiring ground truth depth and poses, thus limiting its applicability.
In contrast, \StructSeg addresses the limitations, being able to process challenging dynamic scenes with moving cameras, even in the absence of depth information and camera poses.

\begin{figure}[t!]
	\centering
	\includegraphics[width=0.999\linewidth]{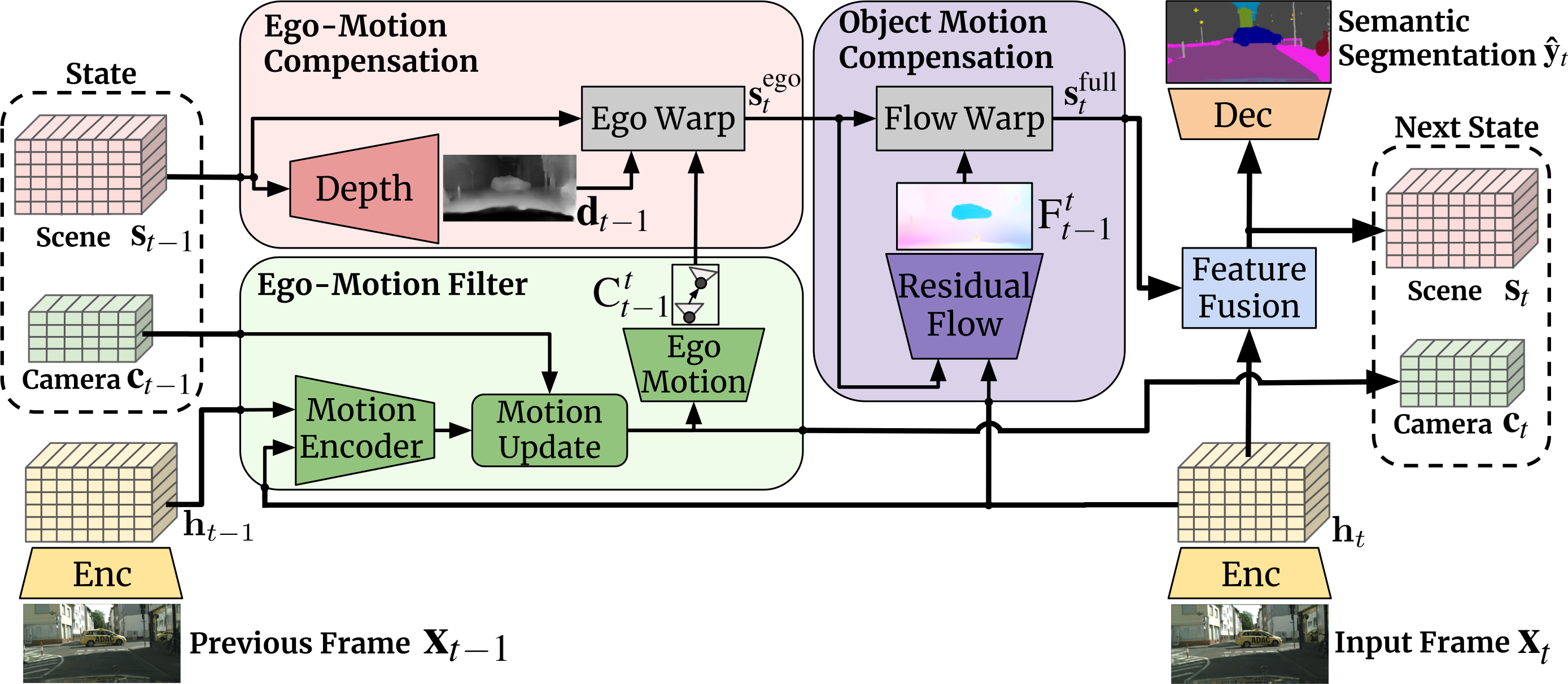}
	\vspace{0.cm}
	\caption{
		\StructSeg structured filter. Scene depth $\Depth_{t-1}$, ego-motion $\CamTFIdx{t-1}{t}$, and object-motion $\FlowIdx{t-1}{t}$ are used to project scene features $\FeatStateIdx{t-1}$ to the current time $t$, where they are fused with current image features $\ImgFeatsIdx{t}$ to obtain a temporally consistent semantic segmentation $\PredSegmentationIdx{t}$.
	}
	\label{fig:main}
\end{figure}

%%%%%%%%%%%%%%%%%%%%%%%%%%%%%%%%%%%%%%%%%%%%%%%%%%%%%%%%%%%%%%%%%%%%%%%%%%%%%%%%%
\section{\StructSeg Structured Filtering Method}

\label{subsec: overview}

We propose \StructSegNosp, illustrated in \Figure{fig:main}, a structured filter that improves the temporal consistency of a semantic segmentation model on moving camera dynamic scene scenarios.
\StructSeg learns in a self-supervised way to estimate geometry and motion representations, i.e., scene depth and camera ego-motion. It also estimates the motion of other agents in the scene, and uses these human-interpretable representations to propagate abstract scene features over time, thus improving its segmentation performance and temporal consistency.

\StructSeg is composed of an image encoder $\ImageEncoder$, a structured filter, and a segmentation decoder~$\SegmentationDecoder$.
It receives as input a sequence of RGB frames $\Images = \{\ImageIdx{1}, ..., \ImageIdx{T}\}$
and encodes them into feature maps \{$\ImgFeatsIdx{1}, ..., \ImgFeatsIdx{T} \}$,
which are then recursively processed to integrate temporal information, and decoded into semantic segmentation maps $\Segmentations = \{ \PredSegmentationIdx{1} ..., \PredSegmentationIdx{T} \}$.

\subsection{Learning of Geometry \& Motion}
\label{subsec: ssl}

\begin{figure}[t!]
	\centering
	\begin{subfigure}[b!]{0.56\textwidth}
		\includegraphics[width=.97\linewidth]{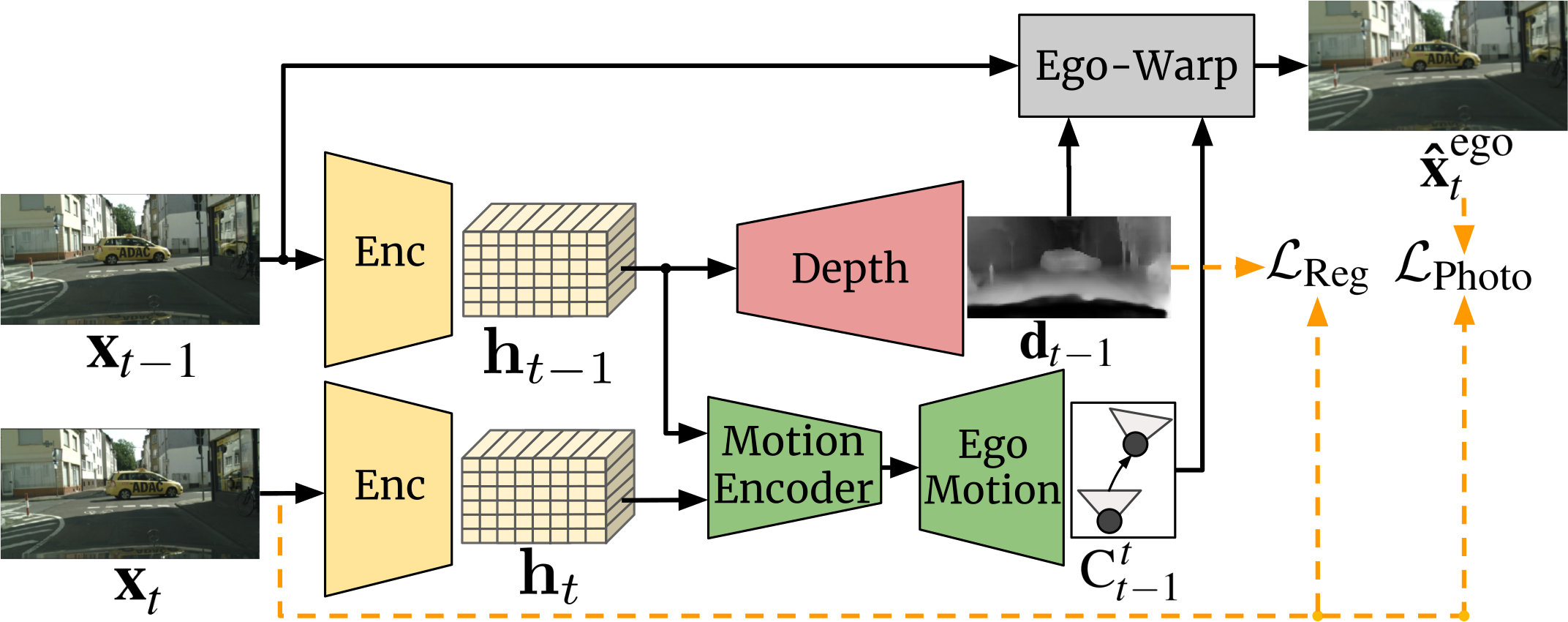}
		\caption{SSL of depth and ego-motion.}
		\label{fig:ssl depth}
	\end{subfigure}
	\begin{subfigure}[b!]{0.43\textwidth}
		\includegraphics[width=0.99\linewidth]{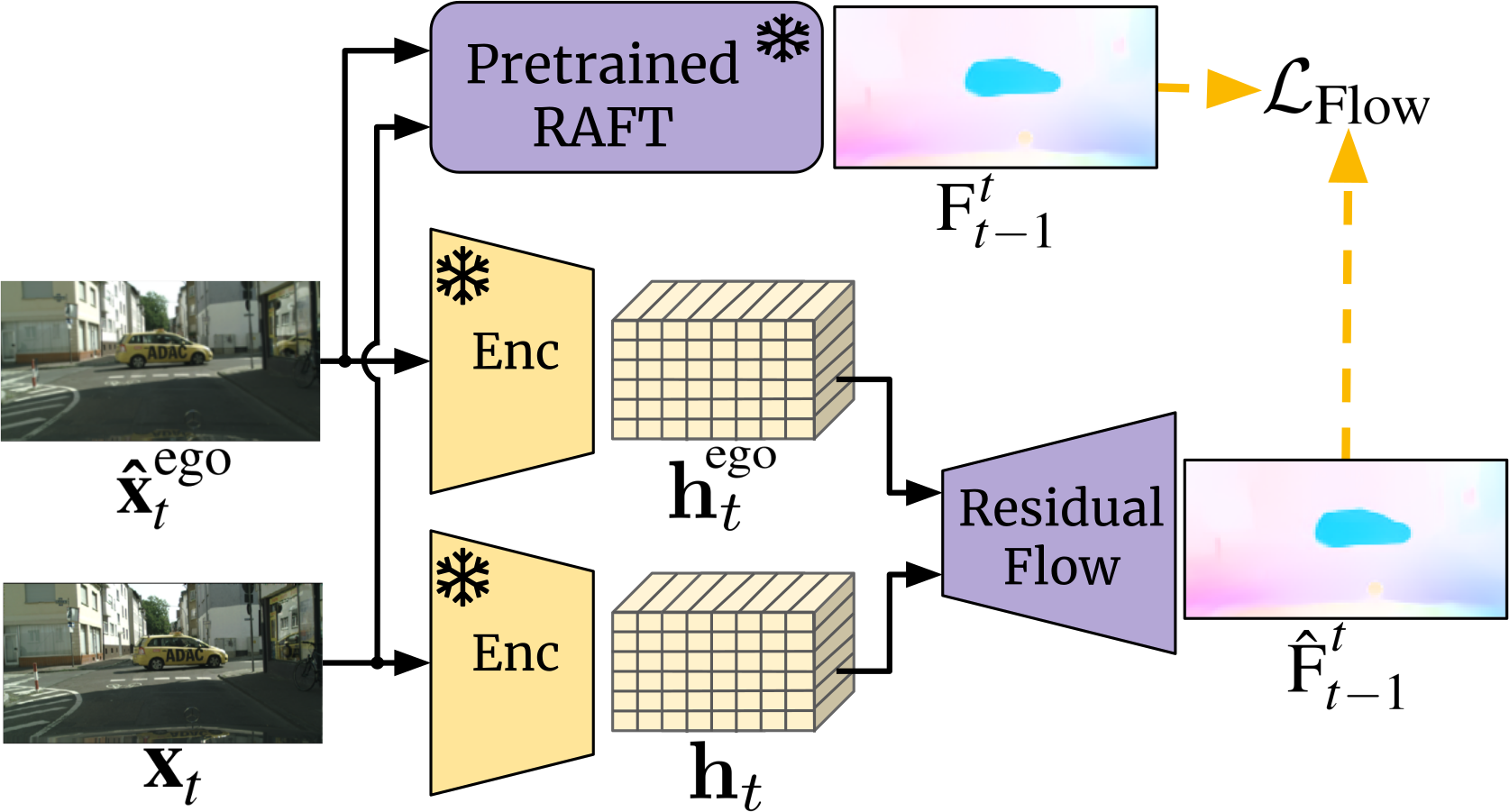}\vspace*{-1ex}
%		\vspace{-0.1cm}
		\caption{Distillation of object-motion.}
		\label{fig:ssl flow}
	\end{subfigure}
	\vspace*{1.5ex}
	\caption{
		Learning geometry and motion.
		\textbf{a)} We learn the scene depth $\Depth_{t-1}$ and ego-motion $\CamTFIdx{t-1}{t}$ in a self-supervised manner given two video frames by enforcing a photometric loss $\LossPhotoEmpty$ between the ego-warped $\EgoImageIdx{t}$ and target frames $\ImageIdx{t}$, as well as a depth regularization $\DepthRegEmpty$.
		\textbf{b)} Given an ego-warped image, we train a residual flow decoder to predict the residual optical flow $\PredFlowIdx{t-1}{t}$ that parameterizes the dynamics of moving objects in the scene by distilling a pretrained RAFT model.}
	\label{fig:ssl}
\end{figure}

\StructSeg learns in a self-supervised manner to estimate scene geometry and camera motion, which are then used to improve the temporal consistency of a segmentation model.
\Figure{fig:ssl} illustrates our two-step self-supervised approach for learning the scene geometry with camera motion and for distillation of object dynamics.

\noindent\textbf{Scene Geometry and Ego-Motion:}
We train our model to predict the monocular depth and camera pose transformation of the vehicle in a self-supervised manner by solving a novel view-synthesis pretext task in  which a target image $\ImageIdx{t}$ is rendered from a source image $\ImageIdx{t-1}$
by modeling the static scene features that change due to the ego-motion~\cite{Zhou_UnsupervisedLearningDethEgoMotionFromVideo_2017, Godard_DiggingIntoSelfSupervisedMonocularDepthEstimation_2019}.

To predict the scene geometry, \StructSeg incorporates a \emph{depth decoder} $\DepthDecoder$ that outputs the depth $\Depth_{t}$ and inverse depth $\InvDepth_{t}$\footnote{
	With slight abuse of notation, we denote the $\InvDepth_{t}$ as \emph{disparities}.
}
of the scene given the input feature maps~$\ImgFeatsIdx{t}$;
whereas to compute the camera motion between two images we employ a \emph{motion encoder} $\MotionEncoder$ that computes motion features between two sets of feature maps, and an \emph{ego-motion decoder} $\CameraPredictor$, which predicts the camera transformation between two time steps $\CamTFIdx{t-1}{t}$, parameterized as a 6-dimensional vector containing the translation parameters and Euler angles of the camera transformation matrix.
We then render the ego-warped image~$\EgoImageIdx{t}$ using the estimated scene depth and ego-motion:
\begin{align}
	& \EgoImageIdx{t} = \ForwardWarp(\ImageIdx{t-1}, \Depth_{t-1}, \CamTFIdx{t-1}{t}, \Intrinsics), \label{eq:warp}
\end{align}
where $\Intrinsics \in \R^{3, 3}$ are the camera intrinsic parameters, and $\ForwardWarp$ is the forward rendering function proposed in~\cite{Lee_LearningMonocularDepthInDynamicScenes_2021}.
If the depth and ego-motion estimates are accurate, the resulting warped images should match the target images except for occluded regions and moving objects.
Therefore, as illustrated in \Figure{fig:ssl depth}, we train the modules for self-supervised depth and ego-motion estimation by optimizing the following loss function:
\begin{align}
	\hspace{-.5em} & \LossDepthEmpty = \LossPhoto{\EgoImageIdx{t}}{\ImageIdx{t}} + \lambda_\textrm{Reg} \cdot \DepthReg{\InvDepth_{t-1}}, \label{eq: geom loss}  \\
	& \LossPhotoEmpty = \frac{\alpha}{2}(1 - \SSIM(\EgoImageIdx{t}, \ImageIdx{t})) + (1 - \alpha) ||\EgoImageIdx{t} - \ImageIdx{t}||_{1}, \hspace{-.5em} \\
	& \DepthRegEmpty = |\partial_x \NormInvDepthIdx{t}| e^{-|\partial_x \ImageIdx{t}|} + |\partial_y \NormInvDepthIdx{t}| e^{-|\partial_y \ImageIdx{t}|},
\end{align}

where $\partial_x$ and $\partial_y$ are the spatial gradients in the $x$- and $y$-directions, $\SSIM$ is the structural similarity index, and $\NormInvDepth$ is the normalized disparity map.
%computed as $\NormInvDepth = \InvDepth / \MeanInvDepth$.
%
$\LossPhotoEmpty$ is a photometric loss that measures the difference between the ego-warped and target images, and $\DepthRegEmpty$ is an edge-aware smoothing regularization~\cite{Godard_UnuspervisedMonocularDepthEstimationWithLeftRightConsistency_2017} that encourages the normalized disparity maps to be locally smooth, except on the image edges.
To mitigate the effect of disocclusions and moving objects during training, we use the auto-masking and per-pixel minimum processing steps proposed in~\cite{Godard_DiggingIntoSelfSupervisedMonocularDepthEstimation_2019}.

\vspace{0.3cm}
\noindent\textbf{Object Motion:} Assuming static scenes as well as accurate depth and ego-motion estimates, the predicted ego-warped images $\EgoImageIdx{t}$ are identical, up to occluded regions, to the target images $\ImageIdx{t}$.
Hence, we make the assumption that any major differences between such frames must be explained by external moving objects (e.g. driving cars or pedestrians).

As illustrated in \Figure{fig:ssl flow}, we estimate the residual optical flow $\PredFlowIdx{t-1}{t}$ between the ego-warped and target images, which encodes the dynamics of moving objects, by training a \emph{residual flow decoder} $\ResidualFlowEncoder$ while keeping all other modules frozen.
The residual flow $\PredFlowIdx{t-1}{t}$ is parameterized as a 2D flow field that encodes the per-pixel motion in the horizontal and vertical directions needed to align the ego-warped images $\EgoImageIdx{t}$ to the corresponding target images $\ImageIdx{t}$.
$\ResidualFlowEncoder$  is trained to match the optical flow predictions of the large state-of-the-art optical flow model RAFT~\cite{Teed_RAFT_2020}:
\vspace{-0.2cm}
\begin{align}
	& \LossFlowEmpty = ||\PredFlowIdx{t-1}{t} - \FlowIdx{t-1}{t}||_1  \label{eq: flow loss}.
\end{align}

\subsection{Structured Filter}
\label{subsec: filter}

The modules and representations described in \Section{subsec: ssl} form the core of the \StructSeg structured filter, which is depicted in \Figure{fig:main}.
It propagates information over time using two different filter states, namely a \emph{scene state}~$\FeatState$ that encodes the scene contents and geometry, and a \emph{camera state}~$\CamState$ that encodes the ego-motion of the vehicle.

It consists of six components: ego-motion filter, depth estimation, ego-motion compensation, residual flow estimation, object motion compensation, and feature fusion.

\noindent\textbf{Ego-Motion Filter:} The ego-motion filter extends the motion encoder $\MotionEncoder$ and ego-motion decoder $\CameraPredictor$ modules in order to aggregate motion information over time and enforce the prediction of temporally consistent ego-motion.
The temporal integration is achieved via a \textit{motion update} module, which is implemented as a ConvGRU~\cite{Ballas_ConvGRU_2015} recurrent layer that jointly processes the motion features and previous camera state $\CamStateIdx{t-1}$, and outputs an updated camera state $\CamStateIdx{t}$ from which the ego-motion can be then predicted:
\begin{align}
	& \CamStateIdx{t} = \CameraUpdate(\MotionEncoder(\ImgFeatsIdx{t}, \ImgFeatsIdx{t-1}), \CamStateIdx{t-1}),
	\hspace{1cm} \CamTFIdx{t-1}{t} = \CameraPredictor(\CamStateIdx{t}).
\end{align}

\noindent\textbf{Depth Estimation and Ego-Motion Compensation:}.
Given the past scene state $\FeatStateIdx{t-1}$, the depth decoder $\DepthDecoder$ computes the depth map $\Depth_{t-1}$. This scene geometry and the estimated ego-motion $\CamTFIdx{t-1}{t}$ are used as in \Equation{eq:warp} to project $\FeatStateIdx{t-1}$ to time $t$.
The resulting ego-warped scene state $\EgoFeatsIdx{t}$ encodes scene contents and geometry after compensation for ego-motion.

\noindent\textbf{Residual Flow Estimation and Object Motion Compensation:} These modules model dynamic objects in the scene, such as pedestrians or vehicles, and update the scene state to compensate for the motion of such objects.
We jointly process the ego-warped scene state $\EgoFeatsIdx{t}$ and the current image features $\ImgFeatsIdx{t}$ with the residual flow decoder $\ResidualFlowEncoder$ in order to compute the residual flow $\PredFlowIdx{t-1}{t}$, which represents the pixel displacement of moving objects between consecutive time steps.
The ego-warped features $\EgoFeatsIdx{t}$ are then projected into the current time-step by applying  the displacement encoded in the residual flow map, followed by bilinear interpolation to obtain valid coordinate values.
The resulting features $\FlowFeatsIdx{t}$ not only incorporate the motion of dynamic objects in the scene, but can also correct alignment errors between $\EgoFeatsIdx{t}$ and $\ImgFeatsIdx{t}$ that might occur due to inaccurate depth or ego-motion estimates.

\noindent\textbf{Feature Fusion:} While the previous modules propagate scene features over time, the \textit{feature fusion} module allows \StructSeg to combine the projected scene features $\FlowFeatsIdx{t}$ with the observed encoded features $\ImgFeatsIdx{t}$.
This fusion operation is performed by an update gate mask $\UpdateGate \in [0, 1]$, which determines in a data-driven manner for each feature map and spatial location whether one can rely on the current features $\ImgFeatsIdx{t}$ or on prior knowledge propagated through the filter $\FlowFeatsIdx{t}$.
The resulting fused scene features $\FeatStateIdx{t}$ are part of the next state of the \StructSeg filter.
More formally, this process can be described as:
\begin{align}
	& \UpdateGate = \sigma(\Conv_{\textrm{s}}(\FlowFeatsIdx{t}) + \Conv_{\textrm{h}}(\ImgFeatsIdx{t}) + \bias,) \\
	& \FeatsIdx{t} = \UpdateGate \odot \FlowFeatsIdx{t} + (1 - \UpdateGate) \odot \ImgFeatsIdx{t},
\end{align}
where $\Conv_{\textrm{s}}$ and $\Conv_{\textrm{h}}$ are convolution blocks, $\bias$ a learned bias, and $\sigma$ the sigmoid function.

\subsection{Model Training}
\label{subsec: training}

\StructSeg consists of multiple components addressing different subtasks: depth estimation, ego-motion estimation, ego-motion compensation, object motion estimation, object motion compensation, and feature fusion.
Naively training such a model in an end-to-end manner with a video segmentation objective can result in bad local optima,
where the model does not learn interpretable representations (e.g. depth or object flow).

To ease the training process, we propose a multi-stage training procedure in which we first train the encoder and decoder modules using image pairs or triplets,
and then integrate and train the filter modules using sequences of six frames in order to gather scene context information and improve the segmentation performance and temporal consistency while retaining interpretable representations.

\begin{table}[t]
	\centering
%	\small
	\footnotesize
	\caption{\StructSeg training stages and hyper-parameters.}
	\label{tab:training}
	\vspace{0.2cm}
	\setlength{\tabcolsep}{6pt}
	\begin{tabular}{P{0.6cm} P{4.2cm} P{3.5cm} P{1.2cm} P{0.9cm}}
		\toprule
		\textbf{Stage} & \textbf{Training Goal} & \textbf{Loss Function} & \textbf{LR} &  \textbf{\# Imgs} \\
		\midrule
		\textbf{1} & Segmentation \& SSL Geometry  & $\LossSegEmpty + \lambda_\textrm{D} \cdot \LossDepthEmpty$ \;  (\ref{eq: geom loss}) & $2 \cdot 10^{-4}$ & 3 \\
		\textbf{2} & Distillation of Object Motion & $\LossFlowEmpty$  \;  (\ref{eq: flow loss}) & $1 \cdot 10^{-4}$ & 2 \\
		\textbf{3} & Ego-Motion Filter & $\LossCamEmpty$  \; (\ref{eq: ego loss}) & $8 \cdot 10^{-5}$ & 6\\
		\textbf{4} & Temporal Integration & $\LossSegEmpty + \lambda_\textrm{TC} \cdot \LossConstEmpty $  \; (\ref{eq: tc loss}) & $8 \cdot 10^{-5}$ & 6 \\
		\bottomrule
	\end{tabular}
	\vspace{-0.0cm}
\end{table}

\StructSeg undergoes a four-stage training process, outlined in~\Table{tab:training}.
Initially, as detailed in \Section{subsec: ssl}, \StructSeg encoder and decoder modules are jointly trained for self-supervised learning of geometry and ego-motion, as well as for semantic segmentation by minimizing a combination of cross entropy $\LossSegEmpty$ and SSL geometry $\LossDepthEmpty$ losses.
Following~\cite{Godard_DiggingIntoSelfSupervisedMonocularDepthEstimation_2019}, we use image triplets $(\ImageIdx{t-\tau}, \ImageIdx{t}, \ImageIdx{t+\tau})$, with $\ImageIdx{t}$ being the target image and $\tau$ being the temporal distance between source and target frames used during this first training stage.
Subsequently, we train the residual flow decoder using image pairs as described in \Section{subsec: ssl} while keeping the remaining modules frozen.
In the third stage, with the goal of improving the temporal continuity of the predicted ego-motion,
we train the ego-motion filter and compensation using short video sequences of length $T$ by minimizing the loss function:
\begin{align}
	& \LossCamEmpty = \frac{1}{T} \sum_{t=1}^{T} \LossPhoto{\EgoImageIdx{t}}{\ImageIdx{t}} , \label{eq: ego loss}
\end{align}
which enforces the model to compute accurate camera motion estimates in order to align the ego-warped state with the current observations.
Finally, in the last training stage we jointly train the feature fusion module and fine-tune the segmentation decoder by minimizing the following loss function:
\begin{align}
	& \LossStructSeg = \frac{1}{T} \sum_{t=1}^{T} \LossSeg{\PredSegmentationIdx{t}}{\SegmentationIdx{t}} + \lambda_\textrm{TC} \cdot \LossConst{\FilterSegmentationIdx{t}}{\PredSegmentationIdx{t}},   \label{eq: tc loss}
\end{align}
where $\LossSegEmpty$ is the cross entropy loss function and $\LossConstEmpty$ is a temporal consistency regularizer that enforces the segmentation $\FilterSegmentationIdx{t}$ computed by decoding $\FlowFeatsIdx{t}$ to be close to the actual predicted segmentation maps $\PredSegmentationIdx{t}$.

%%%%%%%%%%%%%%%%%%%%%%%%%%%%%%%%%%%%%%%%%%%%%%%%%%%%%%%%%%%%%%%%%%%%%%%%%%%%%%%%
\section{Experimental Evaluation}

\subsection{Experiment Setup}
\label{sec: experimental setup}

\noindent\textbf{Dataset:} We evaluate $\StructSeg$ on the Cityscapes~\cite{Cordts_CityscapesDataset_2016} dataset, which contains 5,000 automotive video sequences recorded in 50 German cities. Each sequence contains 30 images of size 1024$\times$2048, where only the 20th frame is annotated.
This dataset is a good benchmark for our model, since it contains real-world dynamic scenes recorded from a moving vehicle.
We augment the data using color jittering, mirroring and random cropping.

\noindent\textbf{Evaluation Metrics:} We evaluate the segmentation performance and temporal consistency of our model. The performance is evaluated using the mean Intersection-over-Union (mIoU).
Following~\cite{Liu_EfficientSemanticVideoSegmentationWithPerFrameInference_2020}, we measure the temporal consistency (TC) of our predicted segmentation maps by computing the mean flow warping error between every two neighboring frames.
Our results are computed using single-scale testing on the full image resolution.

\noindent\textbf{Implementation Details:}
We train two \StructSeg variants using distinct image encoder and segmentation decoder architectures.
Namely, a small variant based on DeepLabV3+~\cite{Chen_DeepLabV3+_2018} with ResNet18~\cite{He_DeepResidualLearningResNet_2016} backbone, and a larger variant based on HRNetV2~\cite{Wang_HRNetVisualRecognition_2020}.
The depth and pose decoders closely follow~\cite{Godard_DiggingIntoSelfSupervisedMonocularDepthEstimation_2019}, which output inverse depth maps and a 6-dimensional vector containing the camera translation and Euler angles, respectively.
Finally, our residual flow decoder is a lightweight version of RAFT~\cite{Teed_RAFT_2020}, for which, to integrate into our filter, we replace the context and feature encoders with a single convolutional block.
We emphasize that \StructSeg is architecture-agnostic and could be implemented with different model designs.
Further implementation details are provided in \Appendix{appendix: details}.

\begin{table}[t!]
	\centering
	\caption
	{
		Comparison of image and video segmentation models on the Cityscapes validation set using small (left) and larger (right) backbones.
		We evaluate the segmentation accuracy (mIoU) and temporal consistency (TC) of the models.
		Best two results are highlighted in boldface and underlined, respectively.
	}
	\label{table: comparison}
	\vspace{0.15cm}
	\footnotesize
	\begin{subtable}{0.49\textwidth}
	\addtolength{\tabcolsep}{-0.2em}
	\begin{tabular}{p{2.2cm}p{1.45cm} p{0.7cm}p{0.7cm}}
		\toprule
		&& \multicolumn{2}{c}{\textbf{Cityscapes}} \\

		\textbf{Model} & \textbf{Backbone} & {\textbf{mIoU}}$\uparrow$ & {\textbf{TC}}$\uparrow$ \\
		\midrule
		DeepLabV3+~\cite{Chen_DeepLabV3+_2018} & ResNet18 & 75.2 & 69.8 \\
		Accel~\cite{Jain_AccelEfficientVideoSemanticSegmentation_2019} & ResNet18 & 72.1 & 70.3 \\
		SKD~\cite{Liu_StructuredKnowldegeDistillationSemanticSegmentation_2019}& ResNet18 & 74.5 & 68.2 \\
		ETC~\cite{Liu_EfficientSemanticVideoSegmentationWithPerFrameInference_2020}& PSPNet18 & 73.1 & 70.6 \\
		ETC~\cite{Liu_EfficientSemanticVideoSegmentationWithPerFrameInference_2020}& MobileNetV2 & 73.9 & 69.9 \\
		TCNet~\cite{Varghese_TC_2021} & ResNet18 & 62.2 & 72.1 \\
		TDNet~\cite{Hu_TemporallyDistributedNetworksForFastVideoSegmentation_2020} & BiSeNet18 & 75.0 & 70.2 \\
		TDNet~\cite{Hu_TemporallyDistributedNetworksForFastVideoSegmentation_2020} & PSPNet18 & \underline{76.8} & 70.4 \\
		STT~\cite{Li_VideoSemanticSegmentationViaSparseTemporalTransformer_2021} & BiSeNet18 & 75.8 & 71.4 \\
		STT~\cite{Li_VideoSemanticSegmentationViaSparseTemporalTransformer_2021} & ResNet18 & \textbf{77.3} & \underline{73.0}\\
		\StructSeg (ours) & ResNet18 & 75.1 & \textbf{74.5} \\
		\bottomrule
	\end{tabular}
	\end{subtable}
	\hfill
	\begin{subtable}{0.49\textwidth}
		\addtolength{\tabcolsep}{-0.12em}
		\vspace{-2.44cm}
		\begin{tabular}{p{2.2cm}p{1.3cm} p{0.7cm}p{0.7cm}}
			\toprule
			&& \multicolumn{2}{c}{\textbf{Cityscapes}} \\

			\textbf{Model} & \textbf{Backbone} & {\textbf{mIoU}}$\uparrow$ & {\textbf{TC}}$\uparrow$ \\
			\midrule
			HRNetV2~\cite{Wang_HRNetVisualRecognition_2020} & HRNetV2 & 76.3 & 70.6 \\
			Accel~\cite{Jain_AccelEfficientVideoSemanticSegmentation_2019} & ResNet50 & 74.2 & - \\
			ETC~\cite{Liu_EfficientSemanticVideoSegmentationWithPerFrameInference_2020}& HRNetV2 & 76.4 & 70.1 \\
			ETC~\cite{Liu_EfficientSemanticVideoSegmentationWithPerFrameInference_2020}& ResNet50 & \textbf{77.9} & 72.3 \\
			AuxAdapt~\cite{Zhang_Auxadapt_2022} & HRNetV2 & 76.6 & \textbf{75.3} \\
			PC~\cite{Zhang_PerceptualConsistencyVideoSegmentation_2022} & HRNetV2  & 76.4 & 71.2 \\
			TCNet~\cite{Varghese_TC_2021} & HRNetV2 & 72.7 & \underline{74.7} \\
			STT~\cite{Li_VideoSemanticSegmentationViaSparseTemporalTransformer_2021} & BiSeNet34 & \underline{77.3} & 72.0 \\
			\StructSeg (ours) & HRNetV2 & {77.1} & {\textbf{75.3}} \\
			\bottomrule
		\end{tabular}
	\end{subtable}
	\vspace{-0.1cm}
\end{table}

\subsection{Comparison with Existing Methods}
\label{sec: quant eval}

%%%%%%%%%%%%%%%%%%%%
% QUANT COMPARISON
%%%%%%%%%%%%%%%%%%%%

In \Table{table: comparison}, we quantitatively compare \StructSeg with several existing image and video segmentation models using small (left) and larger (right) backbones.
For both variants, \StructSeg achieves the highest temporal consistency among all compared methods, while retaining a competitive segmentation performance.
Furthermore, in contrast to other approaches aiming to improve the TC of a segmentation model,
e.g. TCNet~\cite{Varghese_TC_2021},
\StructSeg does not sacrifice segmentation performance in order to improve the temporal consistency,
outperforming multiple VSS models for both backbone variants.
%and even improving mIoU for the HRNetV2 backbone.

\begin{figure}[t!]
	\centering
	\includegraphics[width=0.999\linewidth]{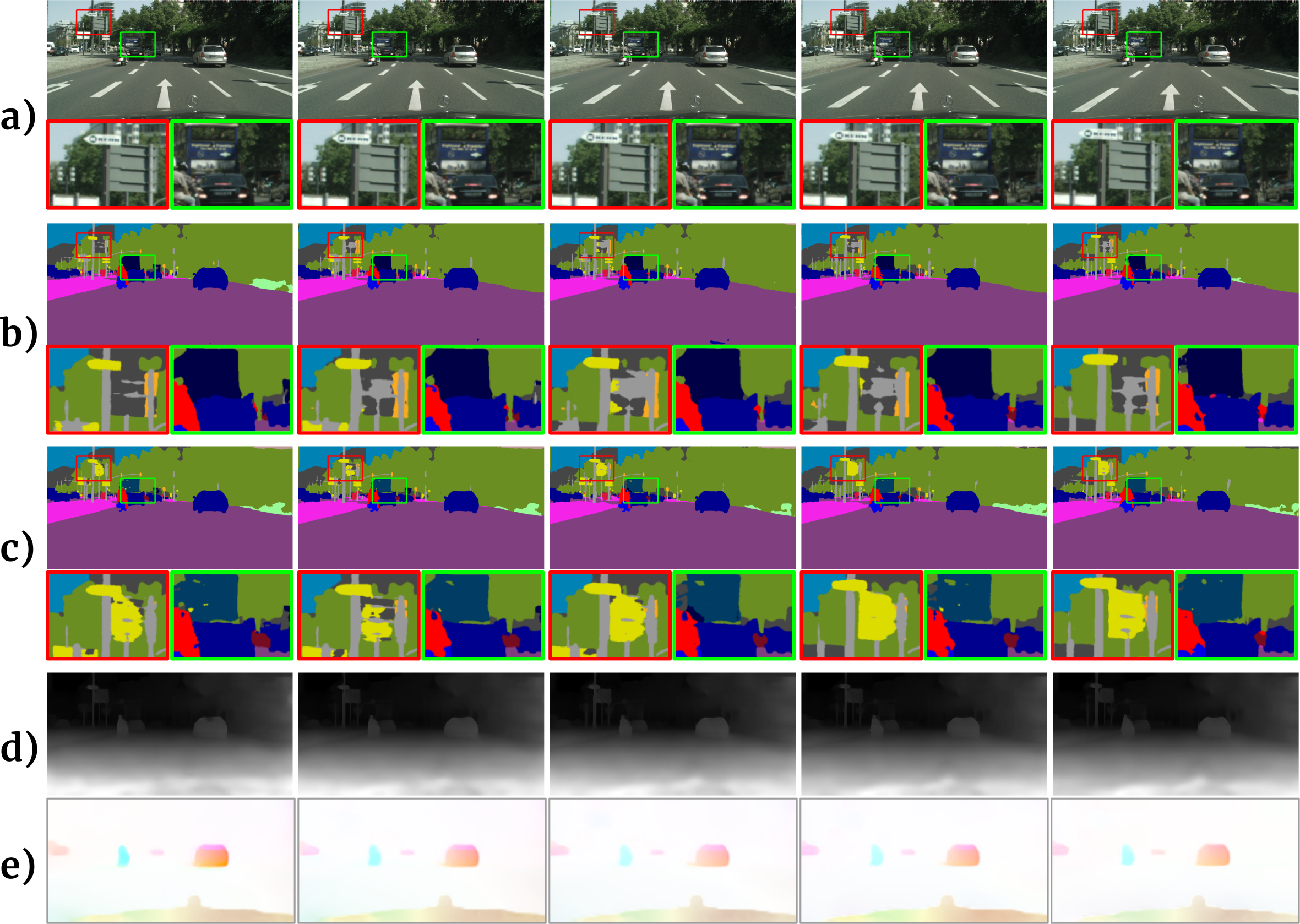}
	\vspace{0.2cm}
	\caption{
		Qualitative evaluation on a validation sequence of five frames.
		\textbf{a)}~Input frames,
		\textbf{b)}~HRNetV2,
		\textbf{c)}~\StructSeg (ours),
		\textbf{d)}~Estimated scene depth,
		\textbf{e)}~Estimated residual flow.
		We highlight areas of the segmentation masks where \StructSeg obtains visibly more accurate and temporally consistent segmentations, such as the traffic signs or the bus, which HRNetV2 mislabels as truck.
	}
	\label{fig:qual eval}
	\vspace{-0.2cm}
\end{figure}

%%%%%%%%%%%%%%%%%%%%
% QUAL COMPARISON
%%%%%%%%%%%%%%%%%%%%
In \Figure{fig:qual eval}, we show a qualitative result comparing \StructSeg with the HRNetV2 baseline on a validation sequence of five frames.
Whereas the baseline mislabels the bus as a truck and outputs inconsistent segmentation labels on certain regions such as the traffic signs, our method achieves more accurate and temporally consistent segmentations, predicting more stable semantic labels across video frames.
Furthermore, we show \StructSeg interpretable intermediate representations, such as the estimated scene depth and residual optical flow, which encodes the movement of the vehicles in the scene, as well as corrections for the hood of the ego-vehicle.
Further visualizations can be found in \Appendix{appendix: qualitative results}.

\subsection{Ablation Study}
\label{sec: ablation study}

To understand the effectiveness of \StructSegNosp, we ablate our filter design and measure the contribution of different steps in our training process.

\noindent\textbf{Filter Design:} Given the same DeepLabV3+ model trained with the SSL procedure described in \Section{subsec: ssl},
we compare \StructSeg with different filter designs, including unstructured RNNs (ConvGRU)~\cite{Siam_ConvolutionalGatedRecurrentNetworksVideoSegmnetation_2017, Pfeuffer_SemanticSegmentaitonVideoSequencesConvLSTM_2019}, optical-flow based filters~\cite{Gadde_SemanticVideoCNNsThroughRepresentationWarping_2017} (flow-only), and a \StructSeg variant modeling only the static scene features (geom-only)~\cite{Wagner_FunctionallyModularSemanticSegmentation_2018}.
The results, reported in \Table{table: filter}, show that filter designs that project scene features using geometry and motion representations outperform the ConvGRU, which learns to model the video dynamics solely from data. 
Furthermore, \StructSegNosp, which decouples the modeling of static and dynamic scene features, achieves the best segmentation performance and temporal consistency among the compared filter designs.

\vspace{0.1cm}
\noindent\textbf{Model Ablation}:
In \Table{table: ablation} we measure the effect that our joint training procedure of semantic segmentation and SSL depth and ego-motion, as well as the \StructSeg filter have on the segmentation performance and temporal consistency.
For two different segmentation models, i.e. DeepLabV3+ with a ResNet18 backbone and HRNetV2, we compare the results after each training stage with those of the model trained for image segmentation only.
First, we note that jointly learning semantic segmentation with SSL depth and ego-motion estimation improves the temporal consistency without significantly compromising the segmentation performance.
We argue that the joint training procedure allows the model to encode the input frames into more robust geometry-aware representations.
Finally, the \StructSeg filter significantly improves the temporal consistency (>4.6\% w.r.t base model), while almost matching the segmentation performance of the base DeepLabV3+ model, and even outperforming HRNetV2.

\begin{figure}[t!]
\begin{minipage}[t!]{0.45\textwidth}
	\centering
	\vspace{-0.5cm}
	\captionof{table}{Comparison of various filter designs. We highlight the diff. to~baseline.}
	\label{table: filter}
	\footnotesize
	\vspace{0.25cm}
	\addtolength{\tabcolsep}{-0.4em}
	\begin{tabular}{p{2.3cm}  p{1.52cm}p{1.52cm}}
		\toprule
		\multicolumn{1}{c}{} & \multicolumn{2}{c}{\hspace{-0.8cm}\textbf{Results}} \\
		\textbf{Model} & \textbf{mIoU}$\uparrow$ & \textbf{TC}$\uparrow$ \\
		\midrule
		ResNet18 + SSL & 74.76 & 70.73 \\
		\hspace{0.1cm} + ConvGRU~\cite{Siam_ConvolutionalGatedRecurrentNetworksVideoSegmnetation_2017} & 73.37 \scriptsize(\textcolor{BrickRed}{\textbf{-1.39}}) & 69.64 \scriptsize(\textcolor{BrickRed}{\textbf{-1.09}}) \\
		\hspace{0.1cm} + Flow-Only~\cite{Gadde_SemanticVideoCNNsThroughRepresentationWarping_2017}  & 74.95 \scriptsize(\textcolor{ForestGreen}{\textbf{+0.19}}) & 73.19 \scriptsize(\textcolor{ForestGreen}{\textbf{+2.46}}) \\
		\hspace{0.1cm} + Geom-Only~\cite{Wagner_FunctionallyModularSemanticSegmentation_2018}  & 74.90 \scriptsize(\textcolor{ForestGreen}{\textbf{+0.14}}) & 73.80 \scriptsize(\textcolor{ForestGreen}{\textbf{+3.07}}) \\
		\hspace{0.1cm} + \StructSeg  & \textbf{75.07} \scriptsize(\textcolor{ForestGreen}{\textbf{+0.31}}) & \textbf{74.53} \scriptsize(\textcolor{ForestGreen}{\textbf{+3.80}}) \\
		\bottomrule
	\end{tabular}
\end{minipage}
\begin{minipage}[t!]{0.02\textwidth}
	\hspace{0.03cm}
\end{minipage}
\begin{minipage}[t!]{0.52\textwidth}
	\centering
	\captionof{table}{Effect of SSL geometry \& motion and \StructSegNosp. We highlight the diff. to~baseline.}
	\label{table: ablation}
	\footnotesize
	\vspace{0.25cm}
	\addtolength{\tabcolsep}{-0.4em}
	\begin{tabular}{p{3.1cm}  p{1.52cm}p{1.45cm}}
		\toprule
		\multicolumn{1}{c}{} & \multicolumn{2}{c}{\hspace{-0.8cm}\textbf{Results}} \\
		\textbf{Model} & \textbf{mIoU}$\uparrow$ & \textbf{TC}$\uparrow$ \\
		\midrule
		ResNet18 &  75.17 & 69.89 \\
		\hspace{0.1cm} + SSL Geom. \& Motion & 74.76 \scriptsize(\textcolor{BrickRed}{\textbf{-0.44}}) & 70.73 \scriptsize(\textcolor{ForestGreen}{\textbf{+0.91}}) \\
		\hspace{0.1cm} + \StructSeg & \textbf{75.07} \scriptsize(\textcolor{BrickRed}{\textbf{-0.10}}) & \textbf{74.53} \scriptsize(\textcolor{ForestGreen}{\textbf{+4.64}}) \\
		\midrule
		HRNetV2 &  76.32 & 70.58 \\
		\hspace{0.1cm} + SSL Geom. \& Motion & 76.45 \scriptsize(\textcolor{ForestGreen}{\textbf{+0.13}}) & 71.65 \scriptsize(\textcolor{ForestGreen}{\textbf{+1.07}}) \\
		\hspace{0.1cm} + \StructSeg & \textbf{77.14} \scriptsize(\textcolor{ForestGreen}{\textbf{+0.82}}) & \textbf{75.34}
		\scriptsize(\textcolor{ForestGreen}{\textbf{+4.76}})\\
		\bottomrule
	\end{tabular}

\end{minipage}
\vspace{-0.5cm}
\end{figure}

%%%%%%%%%%%%%%%%%%%%%%%%%%%%%%%%%%%%%%%%%%%%%%%%%%%%%%%%%%%%%%%%%%%%%%%%%%%%%%%%
\vspace{-0.25cm}
\section{Conclusion}
\vspace{-0.1cm}

We proposed \StructSegNosp, a structured recurrent model for VSS, which learns in a self-supervised manner to estimate scene geometry and camera ego-motion. It also estimates the motion of external objects and leverages these representations to improve the temporal consistency of a semantic segmentation model without sacrificing segmentation performance.
\StructSeg follows a prediction-fusion approach in which scene geometry and camera motion are first used to compensate for ego-motion, then residual flow is used to compensate the motion of dynamic
objects, and finally the projected features are fused with the current observations in order to obtain a temporally consistent representation of the scene.
In our experiments, we showed that \StructSeg outperforms multiple VSS baselines on Cityscapes—achieving superior segmentation temporal consistency and parsing the scene into human-interpretable representations, such as depth, ego-motion and object flow.

%%%%%%%%%%%%%%%%%%%%%%%%%%%%%%%%%%%%%%%%%%%%%%%%%%%%%%%%%%%%%%%%%%%%%%%%%%%%%%%%
\section*{Acknowledgment}
\vspace{-0.1cm}

This work was funded by grant BE 2556/16-2 (Research Unit FOR 2535 Anticipating Human
Behavior) of the German Research Foundation (DFG).

\small
\bibliography{referencesAngel}

\begin{thebibliography}{64}
\providecommand{\natexlab}[1]{#1}
\providecommand{\url}[1]{\texttt{#1}}
\expandafter\ifx\csname urlstyle\endcsname\relax
  \providecommand{\doi}[1]{doi: #1}\else
  \providecommand{\doi}{doi: \begingroup \urlstyle{rm}\Url}\fi

\bibitem[Athar et~al.(2023)Athar, Li, Casas, and Urtasun]{Athar_4DFormer_2023}
Ali Athar, Enxu Li, Sergio Casas, and Raquel Urtasun.
\newblock {4D-Former}: Multimodal {4D} panoptic segmentation.
\newblock In \emph{Conference on Robot Learning (CoRL)}, pages 2151--2164,
  2023.

\bibitem[Baghbaderani et~al.(2024)Baghbaderani, Li, Wang, and
  Qi]{Baghbaderani_TemporallyConsistentVideoSemanticSegmenationBidirectionalOcclusionFeaturePropagation_2024}
Razieh~Kaviani Baghbaderani, Yuanxin Li, Shuangquan Wang, and Hairong Qi.
\newblock Temporally-consistent video semantic segmentation with bidirectional
  occlusion-guided feature propagation.
\newblock In \emph{IEEE/CVF Winter Conference on Applications of Computer
  Vision (WACV)}, pages 685--695, 2024.

\bibitem[Ballas et~al.(2016)Ballas, Yao, Pal, and
  Courville]{Ballas_ConvGRU_2015}
Nicolas Ballas, Li~Yao, Chris Pal, and Aaron Courville.
\newblock Delving deeper into convolutional networks for learning video
  representations.
\newblock \emph{International Conference on Learning Representations (ICLR)},
  2016.

\bibitem[Bian et~al.(2021)Bian, Zhan, Wang, Li, Zhang, Shen, Cheng, and
  Reid]{bian_SCDepth_2021}
Jia-Wang Bian, Huangying Zhan, Naiyan Wang, Zhichao Li, Le~Zhang, Chunhua Shen,
  Ming-Ming Cheng, and Ian Reid.
\newblock Unsupervised scale-consistent depth learning from video.
\newblock \emph{International Journal of Computer Vision (IJCV)}, 129\penalty0
  (9):\penalty0 2548--2564, 2021.

\bibitem[Cao and Behnke(2024)]{Cao_SLCFNet_2024}
Helin Cao and Sven Behnke.
\newblock {SLCF-Net: Sequential LiDAR-Camera Fusion for Semantic Scene
  Completion using a 3D Recurrent U-Net}.
\newblock In \emph{IEEE International Conference on Robotics and Automation
  (ICRA)}, 2024.

\bibitem[Chen et~al.(2018)Chen, Zhu, Papandreou, Schroff, and
  Adam]{Chen_DeepLabV3+_2018}
Liang-Chieh Chen, Yukun Zhu, George Papandreou, Florian Schroff, and Hartwig
  Adam.
\newblock Encoder-decoder with atrous separable convolution for semantic image
  segmentation.
\newblock In \emph{{E}uropean {C}onference on {C}omputer {V}ision (ECCV)},
  pages 801--818, 2018.

\bibitem[Chen et~al.(2021)Chen, Lin, Wang, Yang, and
  Cheng]{Chen_SpatialInformationGuidedConvolutionRGBDSemseg_2021}
Lin-Zhuo Chen, Zheng Lin, Ziqin Wang, Yong-Liang Yang, and Ming-Ming Cheng.
\newblock Spatial information guided convolution for real-time {RGBD} semantic
  segmentation.
\newblock \emph{IEEE Transactions on Image Processing}, 30:\penalty0
  2313--2324, 2021.

\bibitem[Cordts et~al.(2016)Cordts, Omran, Ramos, Rehfeld, Enzweiler, Benenson,
  Franke, Roth, and Schiele]{Cordts_CityscapesDataset_2016}
Marius Cordts, Mohamed Omran, Sebastian Ramos, Timo Rehfeld, Markus Enzweiler,
  Rodrigo Benenson, Uwe Franke, Stefan Roth, and Bernt Schiele.
\newblock The {C}ityscapes dataset for semantic urban scene understanding.
\newblock In \emph{IEEE/CVF Conference on Computer Vision and Pattern
  Recognition (CVPR)}, pages 3213--3223, 2016.

\bibitem[Ding et~al.(2020)Ding, Wang, Zhou, Shi, Lu, and
  Luo]{Ding_EveryFrameCountsLearningVideoSegmentationAndOpticalFlow_2020}
Mingyu Ding, Zhe Wang, Bolei Zhou, Jianping Shi, Zhiwu Lu, and Ping Luo.
\newblock Every frame counts: Joint learning of video segmentation and optical
  flow.
\newblock In \emph{Conference on Artificial Intelligence (AAAI)}, volume~34,
  pages 10713--10720, 2020.

\bibitem[Gadde et~al.(2017)Gadde, Jampani, and
  Gehler]{Gadde_SemanticVideoCNNsThroughRepresentationWarping_2017}
Raghudeep Gadde, Varun Jampani, and Peter~V Gehler.
\newblock Semantic video {CNNs} through representation warping.
\newblock In \emph{IEEE International Conference on Computer Vision (ICCV)},
  2017.

\bibitem[Garg et~al.(2016)Garg, Bg, Carneiro, and
  Reid]{Garg_UnuspervisedCNNSingleViewDepthEstimation_2016}
Ravi Garg, Vijay~Kumar Bg, Gustavo Carneiro, and Ian Reid.
\newblock Unsupervised {CNN} for single view depth estimation: Geometry to the
  rescue.
\newblock In \emph{European Conference on Computer Vision (ECCV)}, pages
  740--756, 2016.

\bibitem[Geiger et~al.(2012)Geiger, Lenz, and Urtasun]{Geiger_KITTI_2012}
Andreas Geiger, Philip Lenz, and Raquel Urtasun.
\newblock Are we ready for autonomous driving? the kitti vision benchmark
  suite.
\newblock In \emph{IEEE/CVF Conference on Computer Vision and Pattern
  Recognition (CVPR)}, pages 3354--3361, 2012.

\bibitem[Godard et~al.(2017)Godard, Mac~Aodha, and
  Brostow]{Godard_UnuspervisedMonocularDepthEstimationWithLeftRightConsistency_2017}
Cl{\'e}ment Godard, Oisin Mac~Aodha, and Gabriel~J Brostow.
\newblock Unsupervised monocular depth estimation with left-right consistency.
\newblock In \emph{IEEE/CVF Conference on Computer Vision and Pattern
  Recognition (CVPR)}, pages 270--279, 2017.

\bibitem[Godard et~al.(2019)Godard, Mac~Aodha, Firman, and
  Brostow]{Godard_DiggingIntoSelfSupervisedMonocularDepthEstimation_2019}
Cl{\'e}ment Godard, Oisin Mac~Aodha, Michael Firman, and Gabriel~J Brostow.
\newblock Digging into self-supervised monocular depth estimation.
\newblock In \emph{IEEE/CVF International Conference on Computer Vision
  (ICCV)}, pages 3828--3838, 2019.

\bibitem[Gordon et~al.(2019)Gordon, Li, Jonschkowski, and
  Angelova]{Gordon_DepthFromVideosInTheWild_2019}
Ariel Gordon, Hanhan Li, Rico Jonschkowski, and Anelia Angelova.
\newblock Depth from videos in the wild: Unsupervised monocular depth learning
  from unknown cameras.
\newblock In \emph{IEEE/CVF International Conference on Computer Vision
  (ICCV)}, pages 8977--8986, 2019.

\bibitem[Guizilini et~al.(2021)Guizilini, Li, Ambrus, and
  Gaidon]{guizilini_geometric_2021}
Vitor Guizilini, Jie Li, Rares Ambrus, and Adrien Gaidon.
\newblock Geometric unsupervised domain adaptation for semantic segmentation.
\newblock In \emph{IEEE/CVF International Conference on Computer Vision
  (ICCV)}, pages 8537--8547, 2021.

\bibitem[He et~al.(2016)He, Zhang, Ren, and
  Sun]{He_DeepResidualLearningResNet_2016}
Kaiming He, Xiangyu Zhang, Shaoqing Ren, and Jian Sun.
\newblock Deep residual learning for image recognition.
\newblock In \emph{IEEE/CVF Conference on Computer Vision and Pattern
  Recognition (CVPR)}, pages 770--778, 2016.

\bibitem[Howard et~al.(2019)Howard, Sandler, Chu, Chen, Chen, Tan, Wang, Zhu,
  Pang, Vasudevan, et~al.]{Howard_MobilenetV3_2019}
Andrew Howard, Mark Sandler, Grace Chu, Liang-Chieh Chen, Bo~Chen, Mingxing
  Tan, Weijun Wang, Yukun Zhu, Ruoming Pang, Vijay Vasudevan, et~al.
\newblock {Searching for MobileNetV3}.
\newblock In \emph{IEEE/CVF International Conference on Computer Vision
  (CVPR)}, pages 1314--1324, 2019.

\bibitem[Hoyer et~al.(2021)Hoyer, Dai, Chen, Koring, Saha, and
  Van~Gool]{hoyer_ThreeWaysToImproveSemanticSegmentation_2021}
Lukas Hoyer, Dengxin Dai, Yuhua Chen, Adrian Koring, Suman Saha, and Luc
  Van~Gool.
\newblock Three ways to improve semantic segmentation with self-supervised
  depth estimation.
\newblock In \emph{IEEE/CVF Conference on Computer Vision and Pattern
  Recognition (CVPR)}, pages 11130--11140, 2021.

\bibitem[Hu et~al.(2020)Hu, Caba, Wang, Lin, Sclaroff, and
  Perazzi]{Hu_TemporallyDistributedNetworksForFastVideoSegmentation_2020}
Ping Hu, Fabian Caba, Oliver Wang, Zhe Lin, Stan Sclaroff, and Federico
  Perazzi.
\newblock Temporally distributed networks for fast video semantic segmentation.
\newblock In \emph{IEEE/CVF Conference on Computer Vision and Pattern
  Recognition (CVPR)}, pages 8818--8827, 2020.

\bibitem[Hur and
  Roth(2016)]{Hur_JointOpticalFlowAndTemporallyConsistentSemSeg_2016}
Junhwa Hur and Stefan Roth.
\newblock Joint optical flow and temporally consistent semantic segmentation.
\newblock In \emph{European Conference on Computer Vision Workshops (ECCVw)},
  pages 163--177. Springer, 2016.

\bibitem[Ilg et~al.(2017)Ilg, Mayer, Saikia, Keuper, Dosovitskiy, and
  Brox]{Ilg_FlowNet2_2017}
Eddy Ilg, Nikolaus Mayer, Tonmoy Saikia, Margret Keuper, Alexey Dosovitskiy,
  and Thomas Brox.
\newblock {Flownet 2.0: Evolution of optical flow estimation with deep
  networks}.
\newblock In \emph{IEEE/CVF Conference on Computer Vision and Pattern
  Recognition (CVPR)}, pages 2462--2470, 2017.

\bibitem[Jain et~al.(2019)Jain, Wang, and
  Gonzalez]{Jain_AccelEfficientVideoSemanticSegmentation_2019}
Samvit Jain, Xin Wang, and Joseph~E Gonzalez.
\newblock Accel: A corrective fusion network for efficient semantic
  segmentation on video.
\newblock In \emph{IEEE/CVF Conference on Computer Vision and Pattern
  Recognition (CVPR)}, pages 8866--8875, 2019.

\bibitem[Kong et~al.(2021)Kong, Shen, and Yang]{Kong_FastFlowNet_2021}
Lingtong Kong, Chunhua Shen, and Jie Yang.
\newblock {FastFlowNet: A lightweight network for fast optical flow
  estimation}.
\newblock In \emph{International Conference on Robotics and Automation (ICRA)},
  pages 10310--10316. IEEE, 2021.

\bibitem[Kundu et~al.(2020)Kundu, Yin, Fathi, Ross, Brewington, Funkhouser, and
  Pantofaru]{Kundu_VirtualMultiViewFusion3DSemanticSegmentation_2020}
Abhijit Kundu, Xiaoqi Yin, Alireza Fathi, David Ross, Brian Brewington, Thomas
  Funkhouser, and Caroline Pantofaru.
\newblock Virtual multi-view fusion for {3D} semantic segmentation.
\newblock In \emph{European Conference on Computer Vision (ECCV)}, pages
  518--535, 2020.

\bibitem[Kuznietsov et~al.(2022)Kuznietsov, Proesmans, and
  Van~Gool]{Kuznietsov_TowardsUnsupervisedOnlineDomainAdaptationForSemanticSegmentation_2022}
Yevhen Kuznietsov, Marc Proesmans, and Luc Van~Gool.
\newblock Towards unsupervised online domain adaptation for semantic
  segmentation.
\newblock In \emph{IEEE/CVF Winter Conference on Applications of Computer
  Vision (WACV)}, pages 261--271, 2022.

\bibitem[Lao et~al.(2023)Lao, Hong, Guo, Zhang, Wang, Chen, and
  Chu]{Lao_SimultaneouslyShortAndLongTermTemporalModelingVSS_2023}
Jiangwei Lao, Weixiang Hong, Xin Guo, Yingying Zhang, Jian Wang, Jingdong Chen,
  and Wei Chu.
\newblock Simultaneously short-and long-term temporal modeling for
  semi-supervised video semantic segmentation.
\newblock In \emph{IEEE/CVF Conference on Computer Vision and Pattern
  Recognition (CVPR)}, pages 14763--14772, 2023.

\bibitem[Lee et~al.(2021)Lee, Im, Lin, and
  Kweon]{Lee_LearningMonocularDepthInDynamicScenes_2021}
Seokju Lee, Sunghoon Im, Stephen Lin, and In~So Kweon.
\newblock Learning monocular depth in dynamic scenes via instance-aware
  projection consistency.
\newblock In \emph{Conference on Artificial Intelligence ({AAAI})}, pages
  1863--1872, 2021.

\bibitem[Li et~al.(2021)Li, Wang, Chen, Niu, Si, Qian, and
  Zhang]{Li_VideoSemanticSegmentationViaSparseTemporalTransformer_2021}
Jiangtong Li, Wentao Wang, Junjie Chen, Li~Niu, Jianlou Si, Chen Qian, and
  Liqing Zhang.
\newblock Video semantic segmentation via sparse temporal transformer.
\newblock In \emph{29th ACM International Conference on Multimedia (MM)}, pages
  59--68, 2021.

\bibitem[Liu et~al.(2019)Liu, Chen, Liu, Qin, Luo, and
  Wang]{Liu_StructuredKnowldegeDistillationSemanticSegmentation_2019}
Yifan Liu, Ke~Chen, Chris Liu, Zengchang Qin, Zhenbo Luo, and Jingdong Wang.
\newblock Structured knowledge distillation for semantic segmentation.
\newblock In \emph{IEEE/CVF Conference on Computer Vision and Pattern
  Recognition (CVPR)}, pages 2604--2613, 2019.

\bibitem[Liu et~al.(2020)Liu, Shen, Yu, and
  Wang]{Liu_EfficientSemanticVideoSegmentationWithPerFrameInference_2020}
Yifan Liu, Chunhua Shen, Changqian Yu, and Jingdong Wang.
\newblock Efficient semantic video segmentation with per-frame inference.
\newblock In \emph{European Conference on Computer Vision (ECCV)}, pages
  352--368, 2020.

\bibitem[Ma et~al.(2017)Ma, St{\"u}ckler, Kerl, and
  Cremers]{Ma_MultiViewDeepLearningSemSegMappingRGBDCameras_2017}
Lingni Ma, J{\"o}rg St{\"u}ckler, Christian Kerl, and Daniel Cremers.
\newblock Multi-view deep learning for consistent semantic mapping with {RGB-D}
  cameras.
\newblock In \emph{IEEE/RSJ International Conference on Intelligent Robots and
  Systems (IROS)}, pages 598--605, 2017.

\bibitem[Novosel et~al.(2019)Novosel, Viswanath, and
  Arsenali]{novosel_BoostingSemanticSegmentationWithMultiTaskSelfSupervisedLearning_2019}
Jelena Novosel, Prashanth Viswanath, and Bruno Arsenali.
\newblock Boosting semantic segmentation with multi-task self-supervised
  learning for autonomous driving applications.
\newblock In \emph{NeurIPS-Workshops}, volume~3, 2019.

\bibitem[Paszke et~al.(2017)Paszke, Gross, Chintala, Chanan, Yang, DeVito, Lin,
  Desmaison, Antiga, and Lerer]{Paszke_AutomaticDifferneciationInPytorch_2017}
Adam Paszke, Sam Gross, Soumith Chintala, Gregory Chanan, Edward Yang, Zachary
  DeVito, Zeming Lin, Alban Desmaison, Luca Antiga, and Adam Lerer.
\newblock Automatic differentiation in {P}y{T}orch.
\newblock In \emph{International Conference on Neural Information Processing
  Systems Workshops ({NeurIPS-W})}, 2017.

\bibitem[Petrovai and Nedevschi(2023)]{Petrovai_MonoDVPSdepthAwareVSS_2023}
Andra Petrovai and Sergiu Nedevschi.
\newblock Monodvps: A self-supervised monocular depth estimation approach to
  depth-aware video panoptic segmentation.
\newblock In \emph{IEEE/CVF Winter Conference on Applications of Computer
  Vision (WACV)}, pages 3077--3086, 2023.

\bibitem[Pfeuffer and
  Dietmayer(2019)]{Pfeuffer_SeparableConvolutionalLSTMFasterVideoSegmentation_2019}
Andreas Pfeuffer and Klaus Dietmayer.
\newblock Separable convolutional lstms for faster video segmentation.
\newblock In \emph{IEEE Intelligent Transportation Systems Conference (ITSC)},
  pages 1072--1078, 2019.

\bibitem[Pfeuffer et~al.(2019)Pfeuffer, Schulz, and
  Dietmayer]{Pfeuffer_SemanticSegmentaitonVideoSequencesConvLSTM_2019}
Andreas Pfeuffer, Karina Schulz, and Klaus Dietmayer.
\newblock Semantic segmentation of video sequences with convolutional lstms.
\newblock In \emph{IEEE Intelligent Vehicles Symposium (IV)}, pages 1441--1447,
  2019.

\bibitem[Qiao et~al.(2021)Qiao, Zhu, Adam, Yuille, and
  Chen]{Qiao_VIPDeepLab_2021}
Siyuan Qiao, Yukun Zhu, Hartwig Adam, Alan Yuille, and Liang-Chieh Chen.
\newblock {ViP-DeepLab: Learning visual perception with depth-aware video
  panoptic segmentation}.
\newblock In \emph{IEEE/CVF Conference on Computer Vision and Pattern
  Recognition (CVPR)}, pages 3997--4008, 2021.

\bibitem[Rosu et~al.(2020)Rosu, Quenzel, and
  Behnke]{Rosu_SemisupervisedSegmanticMappingThroughLabelPropagationWithSemanticTexture_2020}
Radu~Alexandru Rosu, Jan Quenzel, and Sven Behnke.
\newblock Semi-supervised semantic mapping through label propagation with
  semantic texture meshes.
\newblock \emph{International Journal of Computer Vision (IJCV)}, 128\penalty0
  (5):\penalty0 1220--1238, 2020.

\bibitem[Seichter et~al.(2021)Seichter, K{\"o}hler, Lewandowski, Wengefeld, and
  Gross]{Seichter_EfficientRGBDSemanticSegmentationForIndoorSceneAnalysis_2021}
Daniel Seichter, Mona K{\"o}hler, Benjamin Lewandowski, Tim Wengefeld, and
  Horst-Michael Gross.
\newblock Efficient {RGB-D} semantic segmentation for indoor scene analysis.
\newblock In \emph{IEEE International Conference on Robotics and Automation
  (ICRA)}, pages 13525--13531, 2021.

\bibitem[Shelhamer et~al.(2016)Shelhamer, Rakelly, Hoffman, and
  Darrell]{Shelhamer_ClockworkConvnetsForVideoSemanticSegmentation_2016}
Evan Shelhamer, Kate Rakelly, Judy Hoffman, and Trevor Darrell.
\newblock Clockwork convnets for video semantic segmentation.
\newblock In \emph{European Conference on Computer Vision Workshops (ECCVw)},
  pages 852--868, 2016.

\bibitem[Shi et~al.(2020)Shi, Lin, Wang, Hung, and
  Wang]{Shi_SpSequenceNetSegmentation4DPointClouds_2020}
Hanyu Shi, Guosheng Lin, Hao Wang, Tzu-Yi Hung, and Zhenhua Wang.
\newblock {SpSequenceNet: Semantic segmentation network on 4D point clouds}.
\newblock In \emph{IEEE/CVF Conference on Computer Vision and Pattern
  Recognition (CVPR)}, pages 4574--4583, 2020.

\bibitem[Siam et~al.(2017)Siam, Valipour, Jagersand, and
  Ray]{Siam_ConvolutionalGatedRecurrentNetworksVideoSegmnetation_2017}
Mennatullah Siam, Sepehr Valipour, Martin Jagersand, and Nilanjan Ray.
\newblock Convolutional gated recurrent networks for video segmentation.
\newblock In \emph{IEEE International Conference on Image Processing (ICIP)},
  pages 3090--3094, 2017.

\bibitem[Siam et~al.(2018)Siam, Gamal, Abdel-Razek, Yogamani, Jagersand, and
  Zhang]{Siam_ComparativeStudeySegmetationAutonomousDriving_2018}
Mennatullah Siam, Mostafa Gamal, Moemen Abdel-Razek, Senthil Yogamani, Martin
  Jagersand, and Hong Zhang.
\newblock A comparative study of real-time semantic segmentation for autonomous
  driving.
\newblock In \emph{IEEE Conference on Computer Vision and Pattern Recognition
  Workshops (CVPRw)}, pages 587--597, 2018.

\bibitem[Sun et~al.(2022)Sun, Liu, Tang, Chhatkuli, Zhang, and
  Van~Gool]{Sun_MiningRelationsAcrossCrossFrameAffinitiesVSS_2022}
Guolei Sun, Yun Liu, Hao Tang, Ajad Chhatkuli, Le~Zhang, and Luc Van~Gool.
\newblock Mining relations among cross-frame affinities for video semantic
  segmentation.
\newblock In \emph{European Conference on Computer Vision (ECCV)}, pages
  522--539, 2022.

\bibitem[Sun et~al.(2024)Sun, Liu, Ding, Wu, and
  Van~Gool]{Sun_LearningLocalAndGlobalTemporalContextsForVideoSemanticSegmentatio_2024}
Guolei Sun, Yun Liu, Henghui Ding, Min Wu, and Luc Van~Gool.
\newblock Learning local and global temporal contexts for video semantic
  segmentation.
\newblock \emph{IEEE Transactions on Pattern Analysis and Machine Intelligence
  (TPAMI)}, 2024.

\bibitem[Teed and Deng(2020)]{Teed_RAFT_2020}
Zachary Teed and Jia Deng.
\newblock Raft: Recurrent all-pairs field transforms for optical flow.
\newblock In \emph{European Conference on Computer Vision (ECCV)}, pages
  402--419, 2020.

\bibitem[Varghese et~al.(2021)Varghese, Gujamagadi, Klingner, Kapoor, Bar,
  Schneider, Maag, Schlicht, Huger, and Fingscheidt]{Varghese_TC_2021}
Serin Varghese, Sharat Gujamagadi, Marvin Klingner, Nikhil Kapoor, Andreas Bar,
  Jan~David Schneider, Kira Maag, Peter Schlicht, Fabian Huger, and Tim
  Fingscheidt.
\newblock An unsupervised temporal consistency ({TC}) loss to improve the
  performance of semantic segmentation networks.
\newblock In \emph{IEEE/CVF Conference on Computer Vision and Pattern
  Recognition Workshops (CVPRw)}, pages 12--20, 2021.

\bibitem[Villar-Corrales et~al.(2022)Villar-Corrales, Karapetyan, Boltres, and
  Behnke]{Villar_MSPredVideoPrediction_2022}
Angel Villar-Corrales, Ani Karapetyan, Andreas Boltres, and Sven Behnke.
\newblock {MSPred}: Video prediction at multiple spatio-temporal scales with
  hierarchical recurrent networks.
\newblock In \emph{British Machine Vision Conference (BMVC)}, 2022.

\bibitem[Wagner et~al.(2018)Wagner, Fischer, Herman, and
  Behnke]{Wagner_FunctionallyModularSemanticSegmentation_2018}
J{\"o}rg Wagner, Volker Fischer, Michael Herman, and Sven Behnke.
\newblock Functionally modular and interpretable temporal filtering for robust
  segmentation.
\newblock In \emph{British Machine Vision Conference (BMVC)}, page 282, 2018.

\bibitem[Wang et~al.(2018)Wang, Buenaposada, Zhu, and
  Lucey]{Wang_LearningDepthMonoculaVideosDirectMethods_2018}
Chaoyang Wang, Jos{\'e}~Miguel Buenaposada, Rui Zhu, and Simon Lucey.
\newblock Learning depth from monocular videos using direct methods.
\newblock In \emph{IEEE/CVF Conference on Computer Vision and Pattern
  Recognition (CVPR)}, pages 2022--2030, 2018.

\bibitem[Wang et~al.(2021{\natexlab{a}})Wang, Wang, and
  Liu]{Wang_TemporalMemoryAttentionVideoSemanticSegmenation_2021}
Hao Wang, Weining Wang, and Jing Liu.
\newblock Temporal memory attention for video semantic segmentation.
\newblock In \emph{IEEE International Conference on Image Processing (ICIP)},
  pages 2254--2258, 2021{\natexlab{a}}.

\bibitem[Wang et~al.(2020)Wang, Sun, Cheng, Jiang, Deng, Zhao, Liu, Mu, Tan,
  Wang, et~al.]{Wang_HRNetVisualRecognition_2020}
Jingdong Wang, Ke~Sun, Tianheng Cheng, Borui Jiang, Chaorui Deng, Yang Zhao,
  Dong Liu, Yadong Mu, Mingkui Tan, Xinggang Wang, et~al.
\newblock Deep high-resolution representation learning for visual recognition.
\newblock \emph{IEEE Transactions on Pattern Analysis and Machine Intelligence
  (TPAMI)}, 2020.

\bibitem[Wang et~al.(2021{\natexlab{b}})Wang, Dai, Hoyer, Van~Gool, and
  Fink]{Wang_DomainAdaptiveSemanticSegmentationWithSSDE_2021}
Qin Wang, Dengxin Dai, Lukas Hoyer, Luc Van~Gool, and Olga Fink.
\newblock Domain adaptive semantic segmentation with self-supervised depth
  estimation.
\newblock In \emph{IEEE/CVF International Conference on Computer Vision
  (CVPR)}, pages 8515--8525, 2021{\natexlab{b}}.

\bibitem[Xu et~al.(2018)Xu, Fu, Yang, and
  Lee]{Xu_DynamicVideoSegmentationNetwork_2018}
Yu-Syuan Xu, Tsu-Jui Fu, Hsuan-Kung Yang, and Chun-Yi Lee.
\newblock Dynamic video segmentation network.
\newblock In \emph{IEEE/CVF Conference on Computer Vision and Pattern
  Recognition (CVPR)}, pages 6556--6565, 2018.

\bibitem[Yuan et~al.(2022)Yuan, Li, Yang, Cheng, Zhang, Tong, Zhang, and
  Tao]{Yuan_Polyphonicformer_2022}
Haobo Yuan, Xiangtai Li, Yibo Yang, Guangliang Cheng, Jing Zhang, Yunhai Tong,
  Lefei Zhang, and Dacheng Tao.
\newblock Polyphonicformer: Unified query learning for depth-aware video
  panoptic segmentation.
\newblock In \emph{European Conference on Computer Vision (ECCV)}, pages
  582--599. Springer, 2022.

\bibitem[Zama~Ramirez et~al.(2019)Zama~Ramirez, Poggi, Tosi, Mattoccia, and
  Di~Stefano]{Zama_GeometryMeetsSemantics_2019}
Pierluigi Zama~Ramirez, Matteo Poggi, Fabio Tosi, Stefano Mattoccia, and Luigi
  Di~Stefano.
\newblock Geometry meets semantics for semi-supervised monocular depth
  estimation.
\newblock In \emph{14th Asian Conference on Computer Vision (ACCV)}, pages
  298--313. Springer, 2019.

\bibitem[Zhang et~al.(2022{\natexlab{a}})Zhang, Borse, Cai, and
  Porikli]{Zhang_Auxadapt_2022}
Yizhe Zhang, Shubhankar Borse, Hong Cai, and Fatih Porikli.
\newblock {AuxAdapt}: Stable and efficient test-time adaptation for temporally
  consistent video semantic segmentation.
\newblock In \emph{IEEE/CVF Winter Conference on Applications of Computer
  Vision (WACV)}, pages 2339--2348, 2022{\natexlab{a}}.

\bibitem[Zhang et~al.(2022{\natexlab{b}})Zhang, Borse, Cai, Wang, Bi, Jiang,
  and Porikli]{Zhang_PerceptualConsistencyVideoSegmentation_2022}
Yizhe Zhang, Shubhankar Borse, Hong Cai, Ying Wang, Ning Bi, Xiaoyun Jiang, and
  Fatih Porikli.
\newblock Perceptual consistency in video segmentation.
\newblock In \emph{IEEE/CVF Winter Conference on Applications of Computer
  Vision (WACV)}, pages 2564--2573, 2022{\natexlab{b}}.

\bibitem[Zhao et~al.(2017)Zhao, Shi, Qi, Wang, and Jia]{zhao_PSPNet_2017}
Hengshuang Zhao, Jianping Shi, Xiaojuan Qi, Xiaogang Wang, and Jiaya Jia.
\newblock Pyramid scene parsing network.
\newblock In \emph{IEEE Conference on Computer Vision and Pattern Recognition
  (CVPR)}, pages 2881--2890, 2017.

\bibitem[Zhou et~al.(2022)Zhou, Porikli, Crandall, Van~Gool, and
  Wang]{Zhou_SurveyDeepLearningTechniqueVideoSegmentation_2022}
Tianfei Zhou, Fatih Porikli, David~J Crandall, Luc Van~Gool, and Wenguan Wang.
\newblock A survey on deep learning technique for video segmentation.
\newblock \emph{IEEE Transactions on Pattern Analysis and Machine Intelligence
  (TPAMI)}, 45\penalty0 (6):\penalty0 7099--7122, 2022.

\bibitem[Zhou et~al.(2017)Zhou, Brown, Snavely, and
  Lowe]{Zhou_UnsupervisedLearningDethEgoMotionFromVideo_2017}
Tinghui Zhou, Matthew Brown, Noah Snavely, and David~G Lowe.
\newblock Unsupervised learning of depth and ego-motion from video.
\newblock In \emph{IEEE/CVF Conference on Computer Vision and Pattern
  Recognition (CVPR)}, pages 1851--1858, 2017.

\bibitem[Zhu et~al.(2023)Zhu, Han, Cai, Borse, Ghaffari, and
  Porikli]{Zhu_4DPanopticSegmentationInvariantEquivariantFieldPrediction_2023}
Minghan Zhu, Shizhong Han, Hong Cai, Shubhankar Borse, Maani Ghaffari, and
  Fatih Porikli.
\newblock {4D panoptic segmentation as invariant and equivariant field
  prediction}.
\newblock In \emph{IEEE/CVF International Conference on Computer Vision
  (ICCV)}, pages 22488--22498, 2023.

\bibitem[Zhu et~al.(2017)Zhu, Xiong, Dai, Yuan, and
  Wei]{Zhu_DeepFeatureFlowVideoRecognition_2017}
Xizhou Zhu, Yuwen Xiong, Jifeng Dai, Lu~Yuan, and Yichen Wei.
\newblock Deep feature flow for video recognition.
\newblock In \emph{IEEE/CVF Conference on Computer Vision and Pattern
  Recognition (CVPR)}, pages 2349--2358, 2017.

\end{thebibliography}
\normalsize

\newpage

%%%%%%%%%%%%%%%%
%% APPENDICES %%
%%%%%%%%%%%%%%%%
\begin{appendices}

\vspace*{-0.6cm}
\section{Evaluation Metrics}
\label{appendix: eval metrics}

To evaluate \StructSegNosp, we compute its segmentation performance, temporal consistency, throughput and inference speed.

Following the standard practice, we use the mean Intersection over Union (mIoU) to evaluate the segmentation performance.

To evaluate the temporal consistency (TC) of a VSS model, we closely follow the procedure proposed by Liu \etal~\cite{Liu_EfficientSemanticVideoSegmentationWithPerFrameInference_2020}, in which we compute the mean flow warping error between every two neighboring frames.
More precisely, we use FlowNet2~\cite{Ilg_FlowNet2_2017} to compute the optical flow between two adjacent frames, and warp the predicted segmentation maps from time-step $t-1$ into time $t$.
We then calculate the mIoU between the warped and actual target segmentations.
Following~\cite{Liu_EfficientSemanticVideoSegmentationWithPerFrameInference_2020}, we evaluate TC using a subset of 100 sequences from the validation set.

We measure the throughput and inference speed of our model in frames per second (FPS) and milliseconds (ms), respectively.
For this purpose, we perform inference with \StructSeg on 200 different video sequences of 6 frames and average the throughput and time across frames and sequences.

\section{Implementation Details}
\label{appendix: details}

\subsection{Network Details}
\label{appendix: networks}

In this section, we describe the network architectures and operation of each module in \StructSegNosp.
We emphasize that \StructSeg is architecture-agnostic and could be implemented with different, e.g. more powerful or efficient, model designs.

%%%%%%%%%%%%%%%%%%%
%% SEGMENTATION &&%
%%%%%%%%%%%%%%%%%%%

\noindent\textbf{Image Encoder \& Segmentation Decoder:} 
We implement two distinct \StructSeg variants, whose image encoder and segmentation decoder 
% Our image encoder and segmentation decoder modules
follow the architecture of two popular image semantic segmentation models, namely DeepLabV3+~\cite{Chen_DeepLabV3+_2018} with a ResNet18~\cite{He_DeepResidualLearningResNet_2016} backbone and HRNetV2~\cite{Wang_HRNetVisualRecognition_2020} with a channel multiplier of 18, respectively.
In both cases we initialize the parameters of the encoders with those of the model pretrained on ImageNet, whereas the segmentation decoders are initialized with random weights.

%%%%%%%%%%%%%%%%%%%
%% MOTION ENCODER %
%%%%%%%%%%%%%%%%%%%

\noindent\textbf{Motion Encoder:} The motion encoder concatenates the image features from two consecutive time steps (i.e. $\ImgFeatsIdx{t-1}$ and $\ImgFeatsIdx{t}$) across the channel dimension, and processes this representation with three convolutional layers, followed by batch normalization and ReLU activation functions.

\begin{figure}[t!]
	\begin{minipage}[t!]{0.48\textwidth}
		\centering
		\footnotesize
		\captionof{table}{Network architecture of the depth decoder module.}
		\label{tab: depth dec}
		\vspace{0.2cm}
		\setlength{\tabcolsep}{0pt}
		\begin{tabular}{P{1.35cm} P{2.5cm} P{2.3cm}}
			\toprule
			\textbf{Layer} & \textbf{Modules} & \textbf{Output Dim.}\\
			\midrule
			\multirow{2}{*}{Block 1} & Conv + ELU & 256 $\times$ H/8 $\times$ W/8 \\
			& Conv + ELU & 256 $\times$ H/8 $\times$ W/8 \\
			\midrule
			\multirow{2}{*}{Block 2} & Conv + ELU & 128 $\times$ H/8 $\times$ W/8 \\
			& Conv + ELU & 128 $\times$ H/8 $\times$ W/8 \\
			\midrule
			\multirow{2}{*}{Block 3} & Conv + ELU + Ups. & 64 $\times$ H/4 $\times$ W/4 \\
			& Conv + ELU & 64 $\times$ H/4 $\times$ W/4 \\
			\midrule
			\multirow{2}{*}{Block 4} & Conv + ELU + Ups. & 32 $\times$ H/2 $\times$ W/2 \\
			& Conv + ELU & 32 $\times$ H/2 $\times$ W/2 \\
			\midrule
			\multirow{2}{*}{Block 5} & Conv + ELU + Ups. & 16 $\times$ H $\times$ W \\
			& Conv + ELU & 16 $\times$ H $\times$ W \\
			\midrule
			Disp. Pred. & Conv + Sigmoid & 1 $\times$ H $\times$ W \\
			\bottomrule
		\end{tabular}
		\vspace{0.cm}
	\end{minipage}
	\hspace{0.1cm}
	\begin{minipage}[t!]{0.48\textwidth}
		\vspace{-1.845cm}
		\centering
		\footnotesize
		\captionof{table}{Network architecture of the ego-motion decoder module.}
		\label{tab: pose dec}
		\vspace{0.2cm}
		\setlength{\tabcolsep}{0pt}
		\begin{tabular}{P{1.1cm} P{2.7cm} P{2.4cm}}
			\toprule
			\textbf{Layer} & \textbf{Modules} & \textbf{Output Dim.}\\
			\midrule
			{Block 1} & Conv + BN + ReLU & 256 $\times$ H/8 $\times$ W/8 \\
			\midrule
			{Block 2} & Conv + BN + ReLU & 256 $\times$ H/8 $\times$ W/8 \\
			\midrule
			{Block 3} & Conv + BN + ReLU & 128 $\times$ H/8 $\times$ W/8 \\
			\midrule
			{Block 4} & Conv + Pool +  ReLU & 128 $\times$ H/16 $\times$ W/16 \\
			\midrule
			\multirow{2}{*}{\shortstack{Ego \\ Motion}} & Conv & 6 $\times$ H/16 $\times$ W/16 \\
			& Global Avg. Pool & 6 \\
			\bottomrule
		\end{tabular}
		\vspace{0.cm}
	\end{minipage}
\end{figure}

%%%%%%%%%%%%%%%%%%%
%% DEPTH DECODER %%
%%%%%%%%%%%%%%%%%%%

\noindent\textbf{Depth Decoder:} The network architecture of our depth decoder, which is reported in \Table{tab: depth dec}, closely follows~\cite{Godard_UnuspervisedMonocularDepthEstimationWithLeftRightConsistency_2017, Godard_DiggingIntoSelfSupervisedMonocularDepthEstimation_2019}. It is composed of five convolutional blocks using reflection padding, followed by ELU nonlinearities.
Each of the last three convolutional blocks upsamples the feature maps by a factor of two using nearest-neighbor upsampling.
The depth decoder outputs normalized inverse depth maps $\InvDepth$, which are then converted into depth maps $\Depth$ by:
\vspace{-0.0cm}
\begin{align}
	& \frac{1}{\Depth} = \frac{1}{\MinDepth} + (\frac{1}{\MaxDepth} - \frac{1}{\MinDepth}) \cdot \InvDepth \, ,
\end{align}
\vspace{-0.0cm}
where $\MinDepth$ and $\MaxDepth$ are constant values defining the minimum and maximum depth values in the scene, set to $\MinDepth=0.1$m and $\MaxDepth=100$m for the Cityscapes dataset.

%%%%%%%%%%%%%%%%%%%
%% CAM DECODER %%%%
%%%%%%%%%%%%%%%%%%%

\noindent\textbf{Ego-Motion Decoder:} The ego-motion decoder, which is reported in \Table{tab: pose dec}, processes the motion features with a series of convolution layers, followed by batch normalization and ReLU nonlinearities.
The final layer outputs a 6-dimensional vector representing the translation and rotation (parameterized as Euler angles) of the camera transformation matrix.

%%%%%%%%%%%%%%%%%%%
%% FLOW DECODER %%%
%%%%%%%%%%%%%%%%%%%

\noindent\textbf{Residual Flow Decoder:} The residual flow decoder is implemented as a modified lightweight version of RAFT~\cite{Teed_RAFT_2020}.
To seamlessly integrate this module into our \StructSeg filter, 
we modify the implementation of \texttt{raft\_small} provided by PyTorch\footnote{\url{https://pytorch.org/vision/main/models/raft.html}} by replacing the expensive feature and context encoders with a single convolutional block that directly processes the ego-warped $\EgoFeatsIdx{t}$ and image features $\ImgFeatsIdx{t}$.
We set the number of refinement iterations to 12.

%%%%%%%%%%%%%%%%%%%
%% FILTER STUFF %%%
%%%%%%%%%%%%%%%%%%%

\noindent \textbf{\StructSeg Filter:}
We instantiate the \StructSeg filter using the modules described above, after being pretrained for semantic segmentation, SSL of depth and ego-motion, and distillation of object motion.
For the first image in a video sequence, \StructSeg directly predicts its semantic segmentation, without the use of any temporal filtering.
For all other frames, \StructSeg employs the structured filtering method described in the paper.
The scene feature state is initialized with the image features from the first frame in the video sequence ($\FeatStateIdx{1} = \ImgFeatsIdx{1}$), whereas the initial camera state $\CamStateIdx{1}$ is initialized with zeros.
We experimented with learning the initial state representations; however it did not yield any qualitative or quantitative improvements, while increasing the number of learnable parameters.

\subsection{Training and Inference}
\label{appendix: inference}

All our models are implemented in PyTorch~\cite{Paszke_AutomaticDifferneciationInPytorch_2017} and trained with two NVIDIA A100 (80GB) GPUs.
For each of the four training stages undergone by \StructSegNosp, we report in \Table{tab:training_supp} the most relevant hyper-parameters,
including the approximate training time, learning rate, batch size and number of images per sequence.
We empirically set the loss weight values to $\lambda_\textrm{D} = 2 $, $\lambda_\textrm{Reg} = 10^{-5} $ and $\lambda_\textrm{TC} = 1$.

\begin{table}[t!]
	\centering
	\small
	\caption{\StructSeg training stages and hyper-parameters.}
	\label{tab:training_supp}
	\vspace{0.2cm}
	\setlength{\tabcolsep}{0.3pt}
	\begin{tabular}{P{0.75cm} P{4.2cm} P{2.7cm} P{1.1cm} P{1.cm} P{1.8cm} P{1.0cm}}
		\toprule
		\textbf{Stage} & \textbf{Training Goal} & \textbf{Loss Function} & \textbf{LR} & \textbf{Batch} & \textbf{Train Time} & \textbf{\# Imgs} \\
		\midrule
		\textbf{1} & Segmentation \& SSL Geometry  & $\LossSegEmpty + \lambda_\textrm{D} \cdot \LossDepthEmpty$ & $2 \cdot 10^{-4}$ & 12 & 40h & 3 \\
		\textbf{2} & Distillation of Object Motion & $\LossFlowEmpty$ & $1 \cdot 10^{-4}$ & 4 & 18h & 2 \\
		\textbf{3} & Ego-Motion Filter & $\LossCamEmpty$  & $8 \cdot 10^{-5}$  & 8 & 8h & 6 \\
		\textbf{4} & Temporal Integration & $\LossSegEmpty + \lambda_\textrm{TC} \cdot \LossConstEmpty $ & $8 \cdot 10^{-5}$ & 4 & 12h & 6 \\
		\bottomrule
	\end{tabular}
	\vspace{0cm}
\end{table}

\begin{figure}[t!]
	\begin{minipage}[t!]{0.48\textwidth}
		\centering
		\small
		\footnotesize
		\captionof{table}{Throughput, inference speed (in ms) and number of learnable parameters for \StructSeg based on DeepLabV3+.}
		\label{tab: model dl details}
		\vspace{0.2cm}
		\setlength{\tabcolsep}{0pt}
		\begin{tabular}{p{2.7cm} P{1.2cm} P{1.1cm} P{1.1cm}}
			\toprule
			\textbf{Model} & \textbf{\# Params.} & \textbf{FPS}  & \textbf{Inf. (ms)}\\
			\midrule
			Image Enc \, $\ImageEncoder$ & 15.3M & 76.9 & 12.9\\
			Motion Enc \, $\MotionEncoder$ & 4.8M & 279.4  & 3.6 \\
			Motion Update \,  & 2.8M & 476.2 & 2.1 \\
			Ego-Motion Dec \, $\CameraPredictor$ & 1.6M & 492.8 & 2.0 \\
			Depth Dec \, $\DepthDecoder$ & 1.9M & 227.1 & 4.4 \\
			Ego-Motion Comp. & 0 & 63.4 & 15.8 \\
			Residual Flow Dec \, $\ResidualFlowEncoder$ & 2.7M & 14.0 & 71.8 \\
			Object Motion Comp. & 0 & 775.6 & 1.3 \\
			Feature Fusion & 2.6M & 215.3 & 4.6 \\
			Segmentation Dec \ $\SegmentationDecoder$  & 1.3M & 196.7 & 5.1 \\
			\midrule
			Total \StructSeg & 32.9M & 9.0 & 111.6 \\
			\bottomrule
		\end{tabular}
		\vspace{0.cm}
	\end{minipage}
	\hspace{0.1cm}
	\begin{minipage}[t!]{0.48\textwidth}
		\centering
		\small
		\footnotesize
		\captionof{table}{Throughput, inference speed (in ms) and number of learnable parameters for \StructSeg based on HRNetV2.}
		\label{tab: model hrnet details}
		\vspace{0.2cm}
		\setlength{\tabcolsep}{0pt}
		%	\begin{tabular}{p{2.87cm} P{1.35cm} P{1.1cm} P{1.2cm}}
		\begin{tabular}{p{2.7cm} P{1.2cm} P{1.1cm} P{1.1cm}}
			\toprule
			\textbf{Model} & \textbf{\# Params.} & \textbf{FPS} & \textbf{Inf. (ms)}\\
			\midrule
			Image Enc \,  $\ImageEncoder$ & 9.5M & 36.7 & 27.2 \\
			Motion Enc \,  $\MotionEncoder$ & 5.0M & 75.8 & 13.2 \\
			Motion Update \,  & 2.8M & 166.7 & 6.0 \\
			Ego-Motion Dec \,  $\CameraPredictor$ & 1.6M & 174.4 & 5.7 \\
			Depth Dec \,  $\DepthDecoder$ & 1.9M & 62.4 & 16.0 \\
			Ego-Motion Comp. & 0 & 47.9 & 20.9\\
			Residual Flow Dec \,  $\ResidualFlowEncoder$ & 2.9M & 13.4 & 74.6\\
			Object Motion Comp. & 0 & 886.2 & 1.1\\
			Feature Fusion & 2.6M & 76.8 & 13.0 \\
			Segmentation Dec \  $\SegmentationDecoder$  & 78.8K & 627.6 & 1.6 \\
			\midrule
			Total \StructSeg & 26.2M & 5.6 & 177.6 \\
			\bottomrule
		\end{tabular}
		\vspace{0.cm}
	\end{minipage}
\end{figure}

\Tables{tab: model dl details}{tab: model hrnet details} report the number of learnable parameters, throughput and inference time for each individual module, as well as for the complete model, for the \StructSeg variants based on DeepLabV3+ and HRNetV2, respectively.
We emphasize that the ego-motion and object motion compensation modules do not have any learnable parameters, but instead hard-wire our knowledge from the  moving camera dynamic scene domain to project the previous scene features into the current time-step using geometry and motion representations.
We also observe that \StructSeg inference is severely limited by its residual flow decoder. Adapting such module to exploit recent advances in fast optical flow estimation~\cite{Kong_FastFlowNet_2021}), as well as using more efficient image encoders~\cite{Howard_MobilenetV3_2019}, could allow \StructSeg to be used for real time video semantic segmentation.

\section{Quantitative Results}
\label{appendix: quantitative results}

In \Table{tab: results per class} we compare for individual classes of the Cityscapes dataset the segmentation performance and temporal consistency of \StructSeg with an HRNetV2 baseline trained for semantic segmentation and SSL of geometry and motion.
\StructSeg achieves the best segmentation performance and temporal consistency for most classes in the dataset, especially for those corresponding to moving objects, such as \emph{car}, \emph{truck}, \emph{bus} or \emph{train}.

\begin{table}[t!]
	\centering
	\footnotesize
	\caption{
		Segmentation performance (mIoU) and temporal consistency (TC) for individual Cityscapes classes.
		We compare $\StructSeg$	with HRNetV2 backbone with an HRNetV2 model trained for semantic segmentation and SSL of geometry and motion.
	}
	\label{tab: results per class}
	\vspace{0.2cm}
	\setlength{\tabcolsep}{0.18pt}
	\begin{tabular}{p{0.33cm} p{1.15cm}
			p{0.64cm} p{0.64cm} p{0.64cm} p{0.64cm} p{0.64cm} p{0.64cm} p{0.64cm}
			p{0.64cm} p{0.64cm} p{0.64cm} p{0.64cm} p{0.64cm} p{0.64cm} p{0.64cm}
			p{0.64cm} p{0.82cm} p{0.64cm}
		}
		\toprule
		\multicolumn{2}{c}{\textbf{}}  &
		\rotatebox{75}{\textbf{Road}} & \rotatebox{75}{\textbf{Sidewalk}} & \rotatebox{75}{\textbf{Building}} & \rotatebox{75}{\textbf{Wall}} & \rotatebox{75}{\textbf{Fence}} &  \rotatebox{75}{\textbf{Pole}}
		& \rotatebox{75}{\textbf{Traf. Light}} & \rotatebox{75}{\textbf{Traf. Sign}}  & \rotatebox{75}{\textbf{Person}} & \rotatebox{75}{\textbf{Rider}} &  \rotatebox{75}{\textbf{Car}} & \rotatebox{75}{\textbf{Truck}} &  \rotatebox{75}{\textbf{Bus}} & \rotatebox{75}{\textbf{Train}} & \rotatebox{75}{\textbf{Motorbike}} & \rotatebox{75}{\textbf{Bicycle}} & \rotatebox{75}{\textbf{Mean}} \\
		\midrule
		\multirow{2}{*}{\rotatebox{90}{\textbf{mIoU}}}
		& Baseline & \textbf{98.1} & 84.3 & 92.5 & 50.3 & 59.2 & \textbf{66.5} & 70.3 & \textbf{79.3} & \textbf{81.8} & 63.5 & 94.6 & 72.4 & 83.5 & 67.7 & 61.4 & 76.6 & 76.5 \\
		& Ours & \textbf{98.1} & \textbf{84.7} & \textbf{92.7} & \textbf{51.3} & \textbf{59.6} & 65.9 & \textbf{70.5} & \textbf{79.3} & 81.6 & \textbf{63.7} & \textbf{94.9} & \textbf{77.6} &\textbf{85.0} & \textbf{70.6} & \textbf{61.7} & \textbf{76.7} & \textbf{77.1} \\
		\midrule
		\multirow{2}{*}{\rotatebox{90}{\textbf{TC}}}
		& Baseline & 98.7 & 86.0 & 91.1 & 60.8 & 60.1 & 67.1 & 62.4 & 77.1 & 69.0 & 65.8 & 89.0 & 58.1 & 66.8 & 45.8 & 49.4 & 66.6 & 71.7 \\
		& Ours & \textbf{98.9} & \textbf{88.2} & \textbf{92.1} & \textbf{64.1} & \textbf{65.9} & \textbf{69.1} & \textbf{69.2} & \textbf{79.5} & \textbf{70.5} & \textbf{67.8} & \textbf{90.5} &
		\textbf{65.2} & \textbf{69.6} & \textbf{48.3} & \textbf{53.6} & \textbf{71.0} & \textbf{75.3} \\
		\bottomrule
	\end{tabular}
	\vspace{0cm}
\end{table}

\section{Qualitative Results}
\label{appendix: qualitative results}

\subsection{Effect of Each \StructSeg Stage}
\label{appendix: filter}

\begin{figure}[t!]
	\centering
	\includegraphics[width=0.999\linewidth]{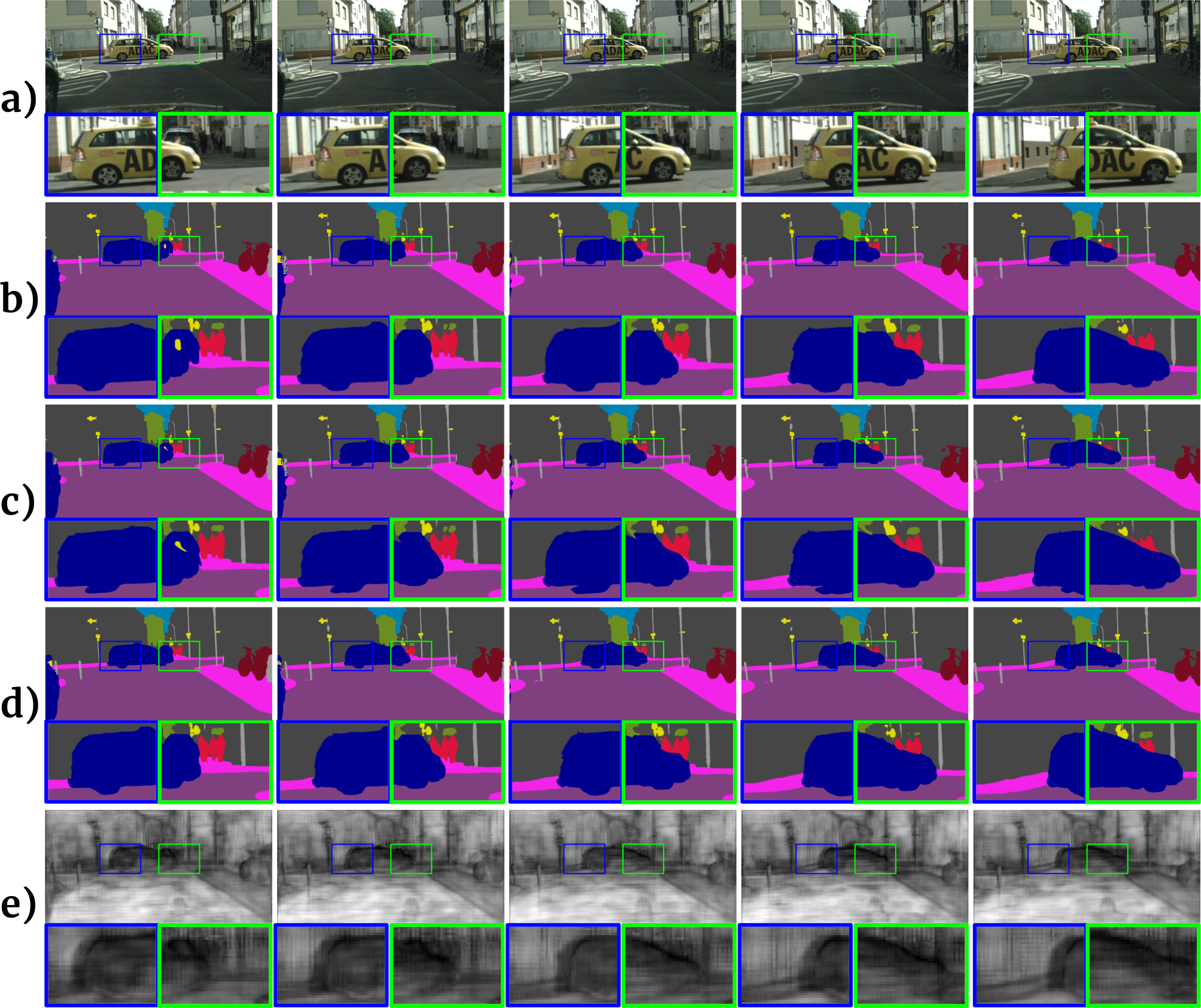}
	\vspace{0.0cm}
	\caption{
		Video segmentation for each stage in \StructSegNosp.
		\textbf{a)}~Input images, 
		\textbf{b)}~segmentation after ego-motion compensation,
		\textbf{c)}~segmentation after object motion compensation,
		\textbf{d)}~segmentation after feature fusion,
		\textbf{e)}~feature fusion update mask, lighter colors mean that filter information is used, whereas darker ones correspond to observations.
	}
	\label{fig:filter stages}
\end{figure}

In \Figure{fig:filter stages} we display the semantic segmentations obtained when decoding the scene features from different stages of our \StructSeg filter.
We can observe how the segmentations after ego-motion compensation (\Figure{fig:filter stages} b) atone for the movement of the ego-vehicle, correctly representing static scene features such as buildings or the bicycle. However, the dynamics of moving objects (e.g. yellow car) are not addressed in this step, thus not compensating for such movement.
This limitation is addressed in the object motion compensation step (\Figure{fig:filter stages} c). 
However, the disocclusions resulting from the moving car and inaccuracies in the residual flow estimation can lead to segmentation errors.
Finally, fusing the projected scene state features with the current observations (\Figure{fig:filter stages} d) leads to a more accurate segmentation of the scene.
In (\Figure{fig:filter stages} e) we visualize the update gate mask $\UpdateGate$ employed for feature fusion. For visualization purposes, we take the mean across all channels and assign lighter colors to spatial locations where \StructSeg relies on the propagated state $\FlowFeatsIdx{t}$, whereas darker colors correspond to the current features $\ImgFeatsIdx{t}$.
We observe that \StructSeg relies on the scene state to represent static areas such as buildings or the street, whereas it relies heavier on observations for accurately segmenting disoccluded areas of the image, fast moving objects or thin structures (e.g. poles or street signs).

\subsection{\StructSeg Qualitative Results}
\label{appendix: qual results}

\begin{figure}[t!]
	\centering
	\includegraphics[width=0.99\linewidth]{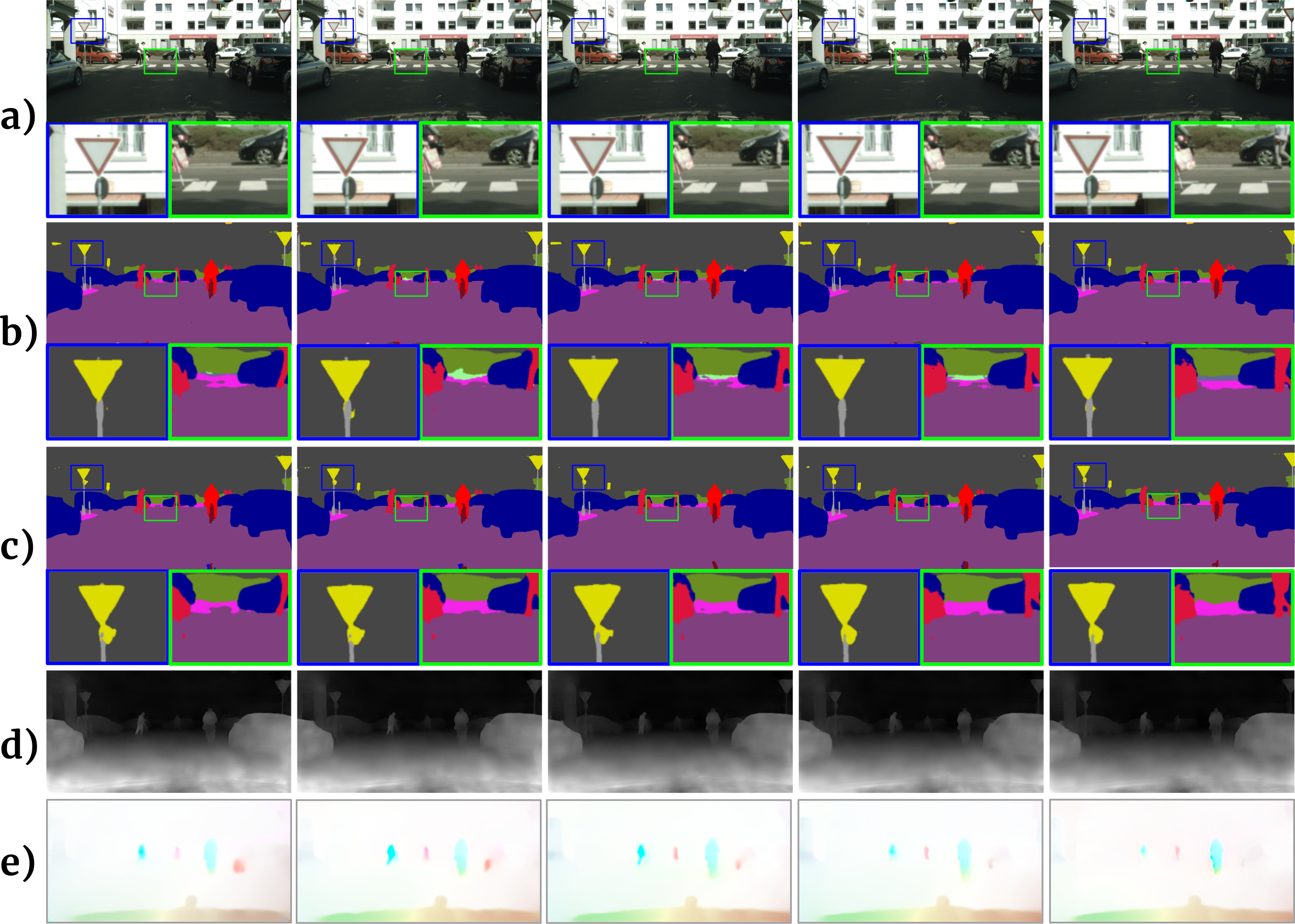}
	\vspace{0.3cm}
	\caption{
		Qualitative evaluation.
		\textbf{a)}~Input frames,
		\textbf{b)}~HRNetV2,
		\textbf{c)}~\StructSeg,
		\textbf{d)}~Estimated scene depth,
		\textbf{e)}~Estimated residual flow.
		We highlight areas of the segmentation masks where \StructSeg obtains visibly more accurate and temporally consistent segmentations.
	}
	\label{fig: qual_comp_1}
\end{figure}

In \Figure{fig: qual_comp_1}, we show a qualitative result comparing \StructSeg with the HRNetV2 baseline.
Our method achieves more accurate and temporally consistent segmentations compared to HRNet, correctly segmenting the traffic signs and reducing the amount of flickering between frames.

\begin{figure}[t!]
	\centering
	\includegraphics[width=0.99\linewidth]{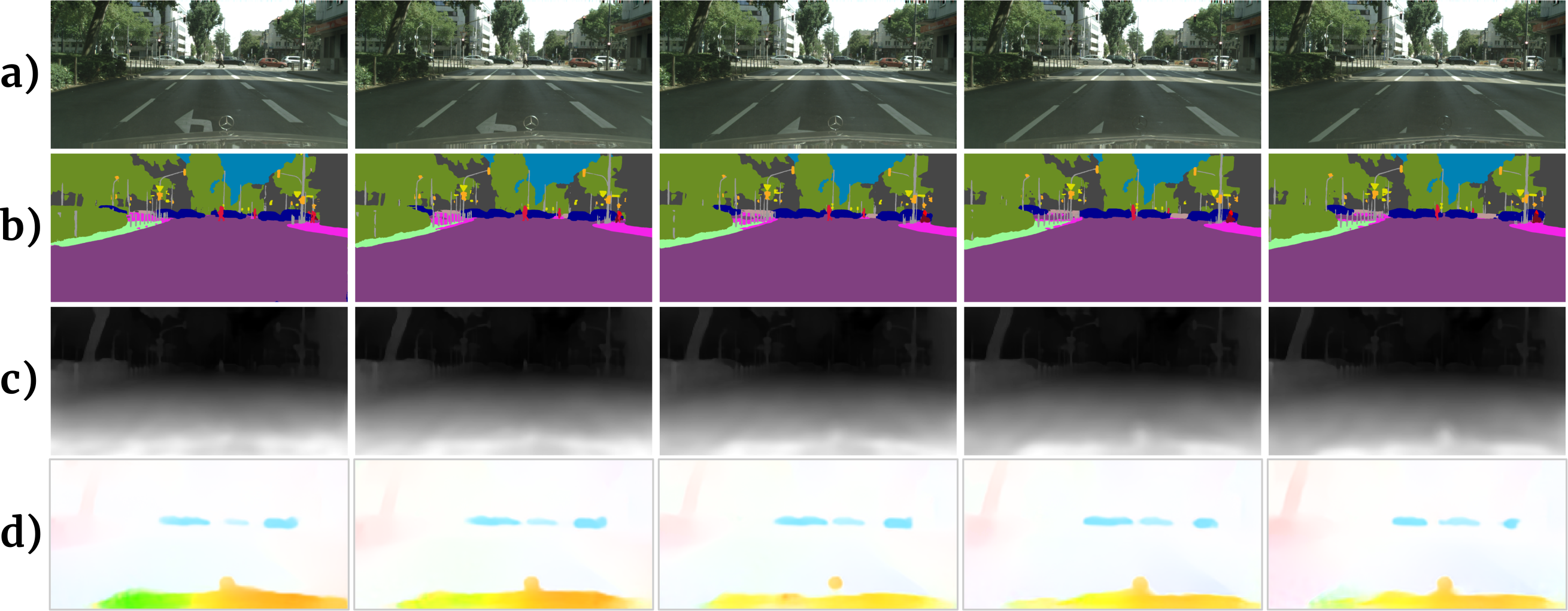}
	\vspace{0.3cm}
	\caption{
		\StructSeg qualitative evaluation.
		\textbf{a)}~Input frames,
		\textbf{b)}~semantic segmentation,
		\textbf{c)}~scene depth,
		\textbf{d)}~residual flow.
	}
	\label{fig: qual_1}
\end{figure}

In \Figuress{fig: qual_1}{fig: qual_4}, we show on four validation sequences how
\StructSeg obtains an accurate and temporally consistent segmentation of the scene, estimates the scene depth, and computes the residual motion flow, which encodes the movement of dynamic objects in the scene, as well as some minor motion corrections for other objects and scene features.

\subsection{Cross-Dataset Evaluation}
\label{appendix: cross dataset}
\vspace{-0.15cm}

\begin{figure}[t!]
	\centering
	\includegraphics[width=0.99\linewidth]{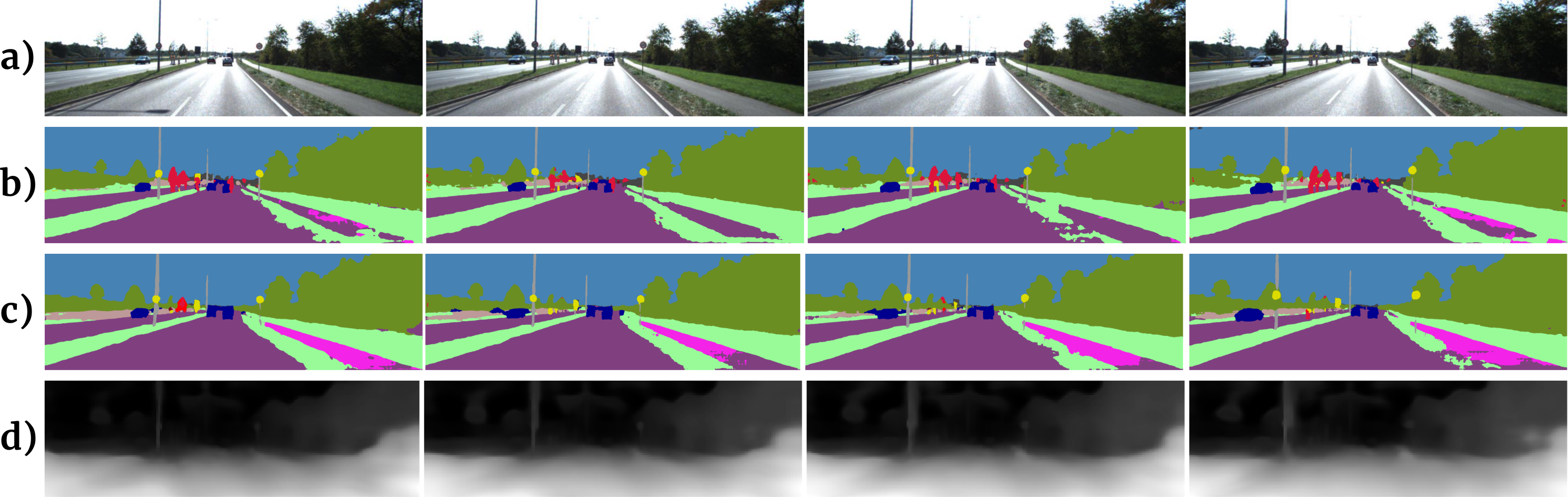}
	\vspace{0.3cm}
	\caption{
		Cross-dataset qualitative evaluation of models trained on Cityscapes and evaluated on KITTI.
		\textbf{a)}~Input frames,
		\textbf{b)}~DeepLabV3+ baseline,
		\textbf{c)}~\StructSeg (ours),
		\textbf{d)}~estimated scene depth.
	}
	\label{fig: cross_db_1}
\end{figure}

We qualitatively evaluate the robustness of \StructSeg by performing a cross-dataset validation in which a DeepLabV3+ baseline and our \StructSeg model trained on Cityscapes are qualitatively evaluated without retraining on sequences from the KITTI~\cite{Geiger_KITTI_2012} dataset.
\Figures{fig: cross_db_1}{fig: cross_db_2}illustrate the semantic segmentation predictions of both models, as well as the \StructSeg depth estimates on two validation sequences of the KITTI dataset.
Due to differences with respect to the training data in the camera model and calibration, as well as different image resolution and aspect ratio, the segmentation performance of both models on the KITTI dataset is severely degraded with respect to Cityscapes.
However, $\StructSeg$ achieves a more accurate and temporally consistent video segmentation, thus verifying that incorporating geometry and motion inductive biases from the moving camera dynamic scene domain into the VSS model design leads to more robust representations and segmentation results.

\subsection{Point Clouds}
\label{appendix: point clouds}
\vspace{-0.15cm}

\begin{figure}[t!]
	\centering
	\includegraphics[width=.99\linewidth]{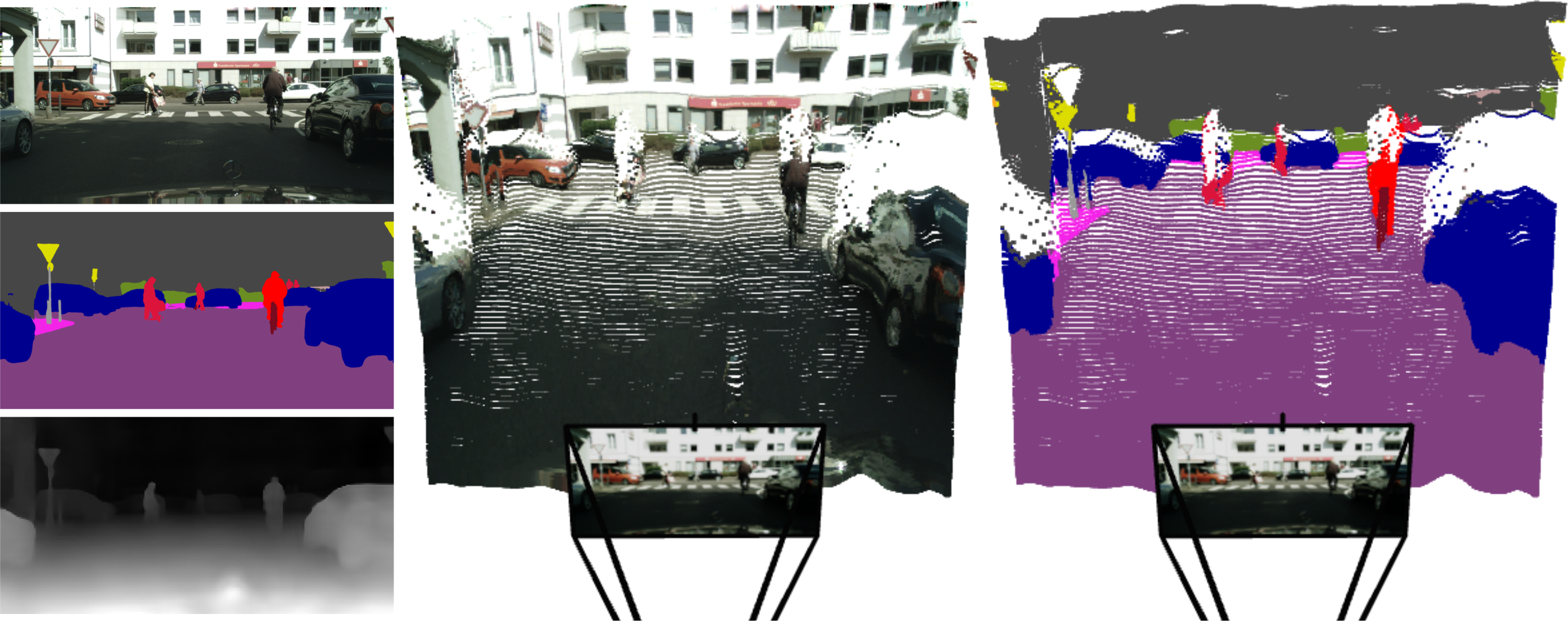}
	\vspace*{1ex}
	\caption{RGB and semantic point clouds rendered by lifting image values and semantic labels to 3D space.}
	\label{fig: point clouds 1}
\end{figure}

\Figuress{fig: point clouds 1}{fig: point clouds 4} show examples of RGB and semantic point clouds rendered by backprojecting image values and semantic labels using the depth maps estimated by \StructSeg and known camera intrinsics.
The high-quality depth maps computed by \StructSeg allow for an accurate 3D representation of the scene.

%%%%%%%%%%%%%%%%%%%%%%%
% QUALITATIVE RESULTS %
%%%%%%%%%%%%%%%%%%%%%%%

\begin{figure}[t!]
	\centering
	\includegraphics[width=0.99\linewidth]{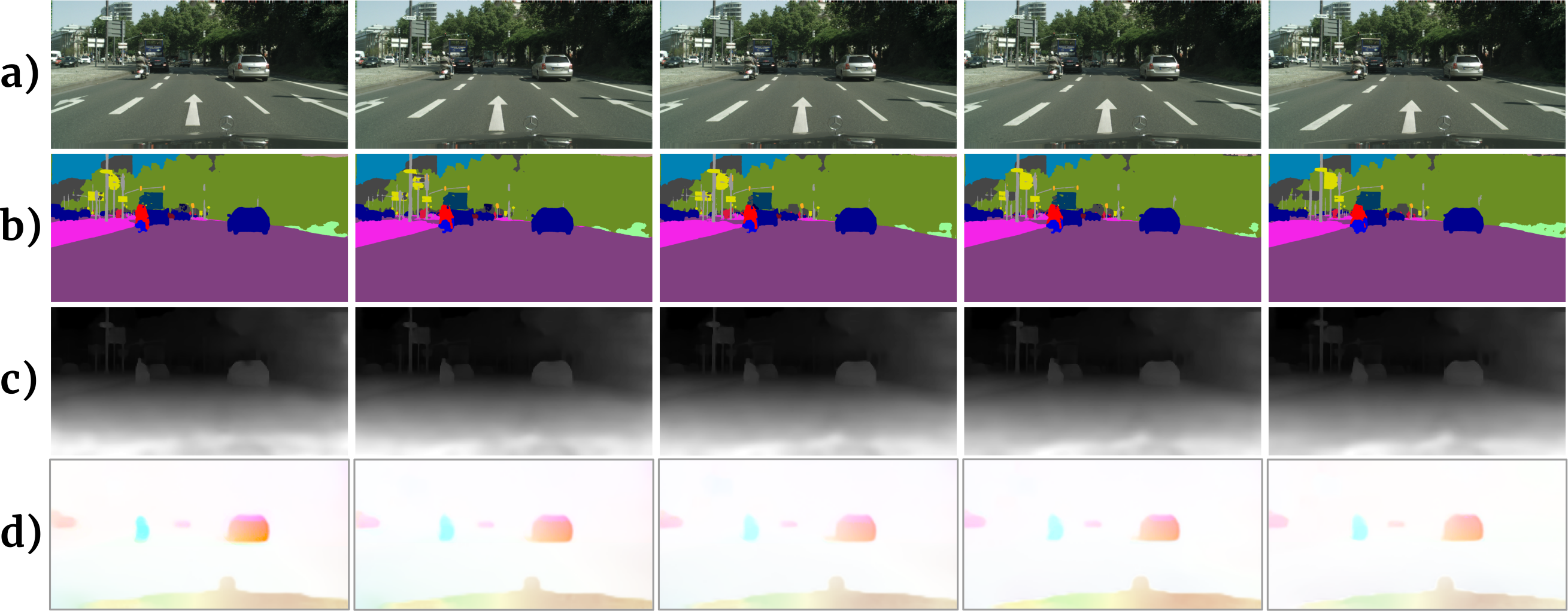}
	\vspace{0.3cm}
	\caption{
		\StructSeg qualitative evaluation.
		\textbf{a)}~Input frames,
		\textbf{b)}~semantic segmentation,
		\textbf{c)}~scene depth,
		\textbf{d)}~residual flow.
	}
	\label{fig: qual_2}
\end{figure}

\begin{figure}[t!]
	\centering
	\includegraphics[width=0.99\linewidth]{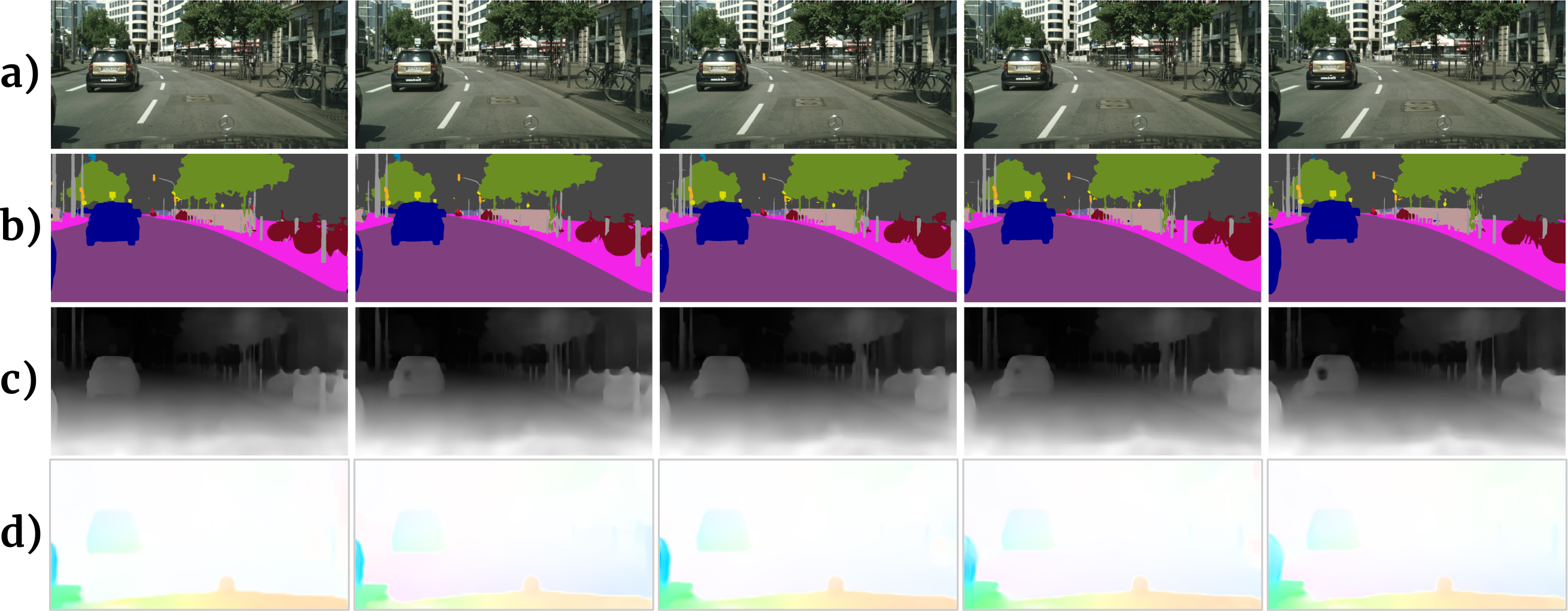}
	\vspace{0.3cm}
	\caption{
		\StructSeg qualitative evaluation.
		\textbf{a)}~Input frames,
		\textbf{b)}~semantic segmentation,
		\textbf{c)}~scene depth,
		\textbf{d)}~residual flow.
	}
	\label{fig: qual_3}
\end{figure}

\begin{figure}[t!]
	\centering
	\includegraphics[width=0.99\linewidth]{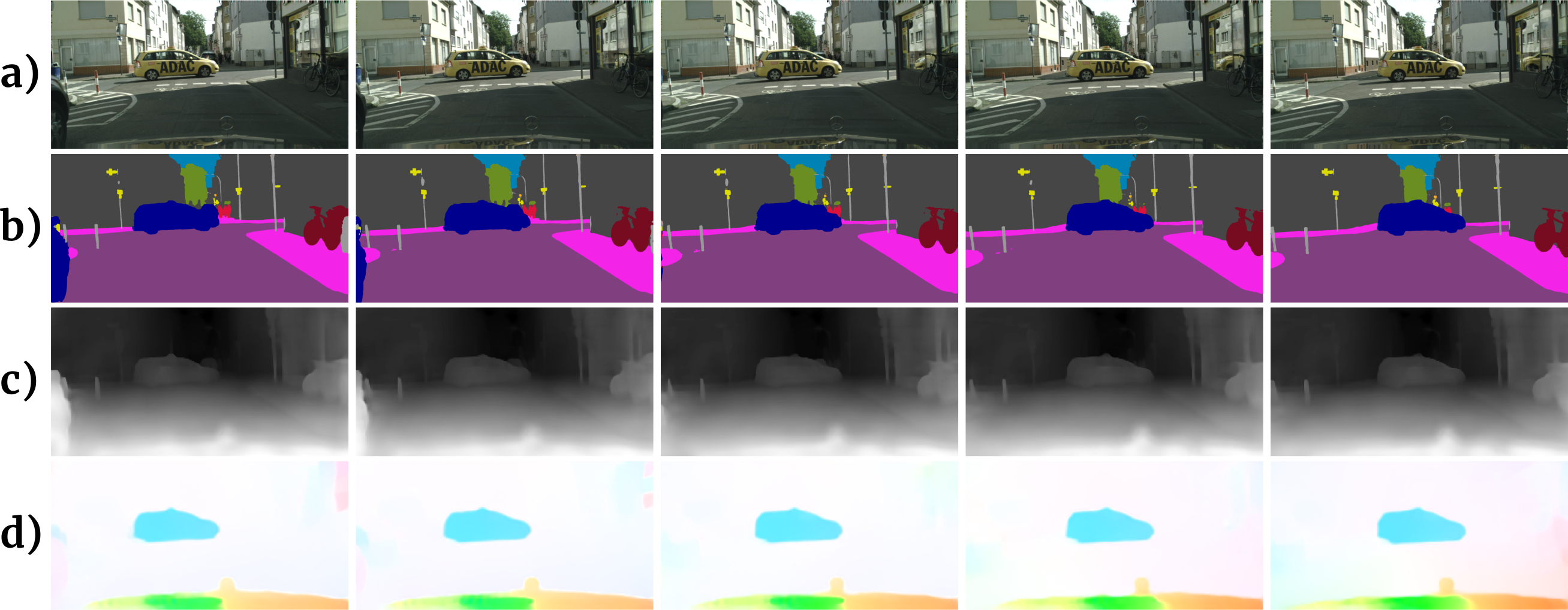}
	\vspace{0.3cm}
	\caption{
		\StructSeg qualitative evaluation.
		\textbf{a)}~Input frames,
		\textbf{b)}~semantic segmentation,
		\textbf{c)}~scene depth,
		\textbf{d)}~residual flow.
	}
	\label{fig: qual_4}
\end{figure}

%%%%%%%%%%%%%%%%%
% CROSS-DATASET %
%%%%%%%%%%%%%%%%%

\begin{figure}[t!]
	\centering
	\includegraphics[width=0.99\linewidth]{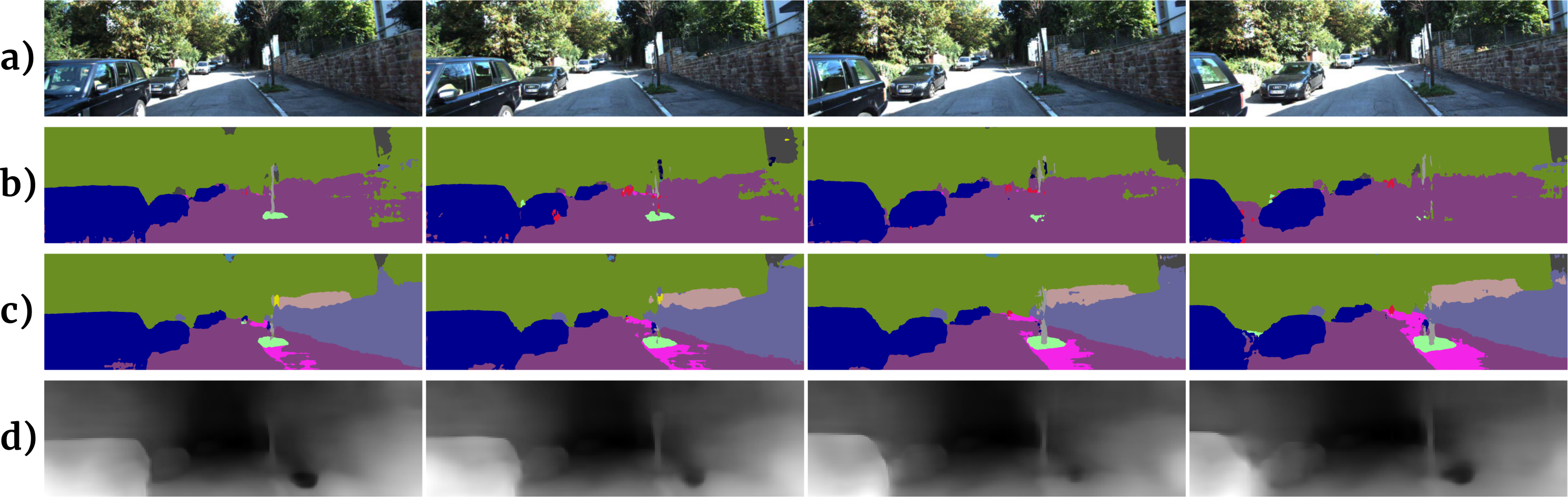}
	\vspace{0.3cm}
	\caption{
		Cross-dataset qualitative evaluation of models trained on Cityscapes and evaluated on KITTI.
		\textbf{a)}~Input frames,
		\textbf{b)}~DeepLabV3+ baseline,
		\textbf{c)}~\StructSeg (ours),
		\textbf{d)}~estimated scene depth.
	}
	\label{fig: cross_db_2}
\end{figure}

%%%%%%%%%%%%%%%%
% POINT CLOUDS %
%%%%%%%%%%%%%%%%

\begin{figure}[t!]
	\centering
	\includegraphics[width=.99\linewidth]{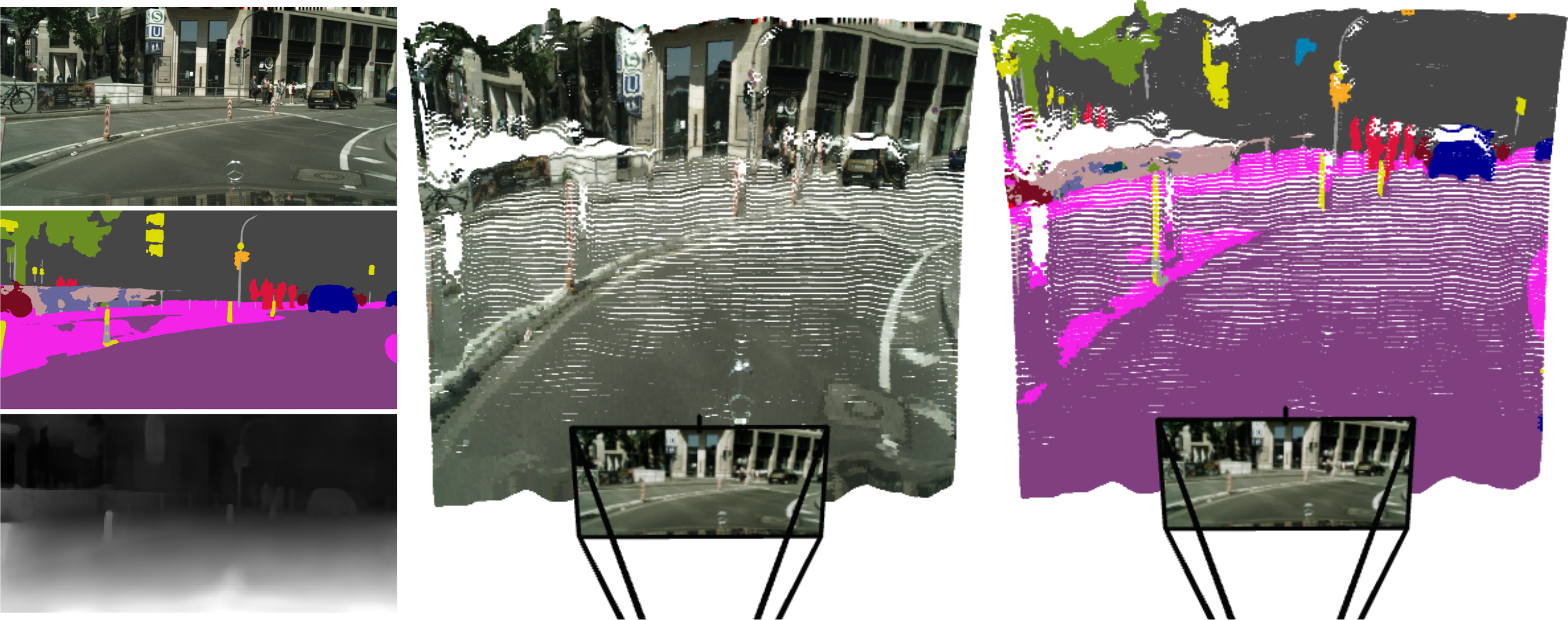}
	\vspace*{1ex}
	\caption{RGB and semantic point clouds rendered by lifting image values and semantic labels to 3D space.}
	\label{fig: point clouds 2}
\end{figure}

\begin{figure}[t!]
	\centering
	\includegraphics[width=.99\linewidth]{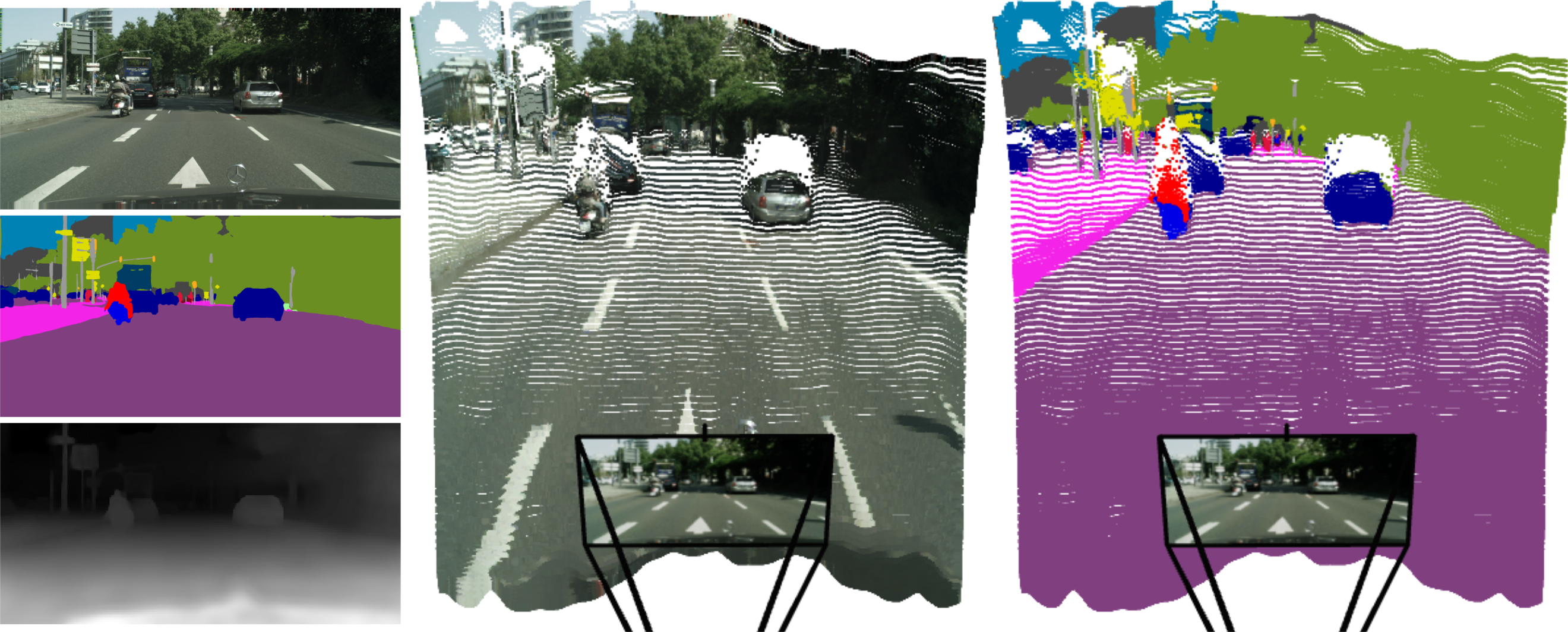}
	\vspace*{1ex}
	\caption{RGB and semantic point clouds rendered by lifting image values and semantic labels to 3D space.}
	\label{fig: point clouds 3}
\end{figure}

\begin{figure}[t!]
	\centering
	\includegraphics[width=.99\linewidth]{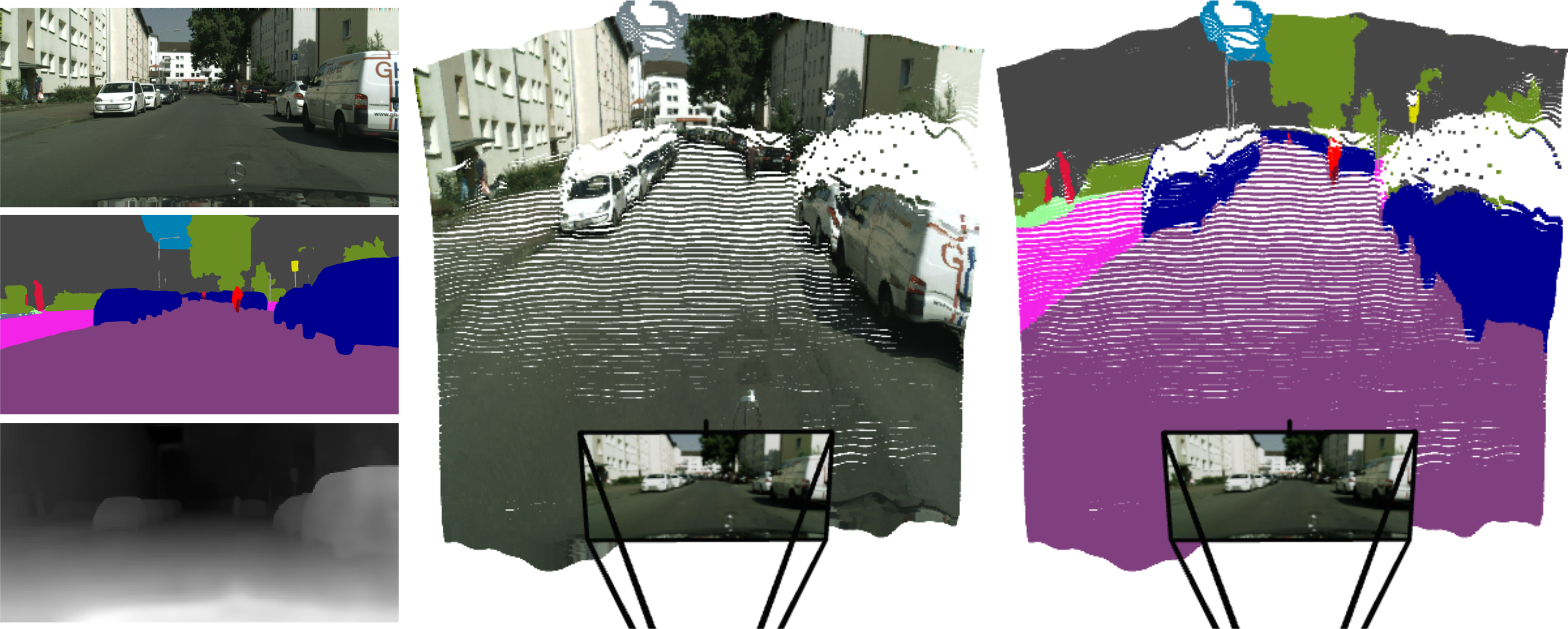}
	\vspace*{1ex}
	\caption{RGB and semantic point clouds rendered by lifting image values and semantic labels to 3D space.}
	\label{fig: point clouds 4}
\end{figure}

\end{appendices}

\end{document}